%% file: ms.tex
\documentclass[10pt,twocolumn,letterpaper]{article}
\pdfoutput=1

\usepackage{cvpr}
\usepackage{times}
\usepackage{epsfig}
\usepackage{graphicx}
\usepackage{amsmath}
\usepackage{amssymb}
\usepackage{bm}
\usepackage[linesnumbered,ruled]{algorithm2e}
\usepackage[utf8]{inputenc}

\usepackage{mynotation}
\usepackage[pagebackref=true,breaklinks=true,letterpaper=true,colorlinks,bookmarks=false]{hyperref}

\cvprfinalcopy 

\newcommand{\parsection}[1]{\noindent\textbf{#1:} }

\def\cvprPaperID{} 

\ifcvprfinal\fi
\begin{document}

\title{A Generative Appearance Model for End-to-end Video Object Segmentation}

\author{Joakim Johnander$^{1,4}$\\
\and
Martin Danelljan$^{1}$\\
\and
Emil Brissman$^{1,3}$\\
\and
Fahad Shahbaz Khan$^{1,2}$\\
\and
Michael Felsberg$^{1}$\\
\\
$^1$ Computer Vision Laboratory, Department of Electrical Engineering, Linköping University\\
$^2$ Inception Institute of Artificial Intelligence, Abu Dhabi, UAE\\
$^3$ Saab, Sweden\\
$^4$ Zenuity, Sweden\\
{\tt\small \{joakim.johnander,martin.danelljan,emil.brissman,fahad.khan,michael.felsberg\}@liu.se}\\
}

\maketitle

\begin{abstract}
  \input{abstract}

\end{abstract}

\input{introduction}

\input{relatedwork}

\input{method}

\input{experiments}

\input{conclusion}

{\small
\bibliographystyle{ieee}
\bibliography{references}
}

\end{document}

%% file: abstract.tex
One of the fundamental challenges in video object segmentation is to find an effective representation of the target and background appearance. The best performing approaches resort to extensive fine-tuning of a convolutional neural network for this purpose. Besides being prohibitively expensive, this strategy cannot be truly trained end-to-end since the online fine-tuning procedure is not integrated into the offline training of the network.

To address these issues, we propose a network architecture that learns a powerful representation of the target and background appearance in a single forward pass. The introduced appearance module learns a probabilistic generative model of target and background feature distributions. Given a new image, it predicts the posterior class probabilities, providing a highly discriminative cue, which is processed in later network modules. Both the learning and prediction stages of our appearance module are fully differentiable, enabling true end-to-end training of the entire segmentation pipeline. Comprehensive experiments demonstrate the effectiveness of the proposed approach on three video object segmentation benchmarks. We close the gap to approaches based on online fine-tuning on DAVIS17, while operating at 15 FPS on a single GPU. Furthermore, our method outperforms all published approaches on the large-scale YouTube-VOS dataset.

%% file: introduction.tex
\section{Introduction}
Video object segmentation (VOS) is the task of tracking \emph{and} segmenting one or multiple target objects in a video sequence. In this work, we consider the semi-supervised setting, where the ground-truth segmentation is only given in the first frame. The task is generic, i.e., the targets are arbitrary and no further assumptions regarding the object classes are made. The VOS problem is challenging from several aspects. The target may undergo significant appearance changes and may be subject to fast motion or occlusion. Moreover, the scene may contain distractor objects that are visually or semantically similar to the target.

\begin{figure}[!t]  
  \includegraphics[width=.32\columnwidth]{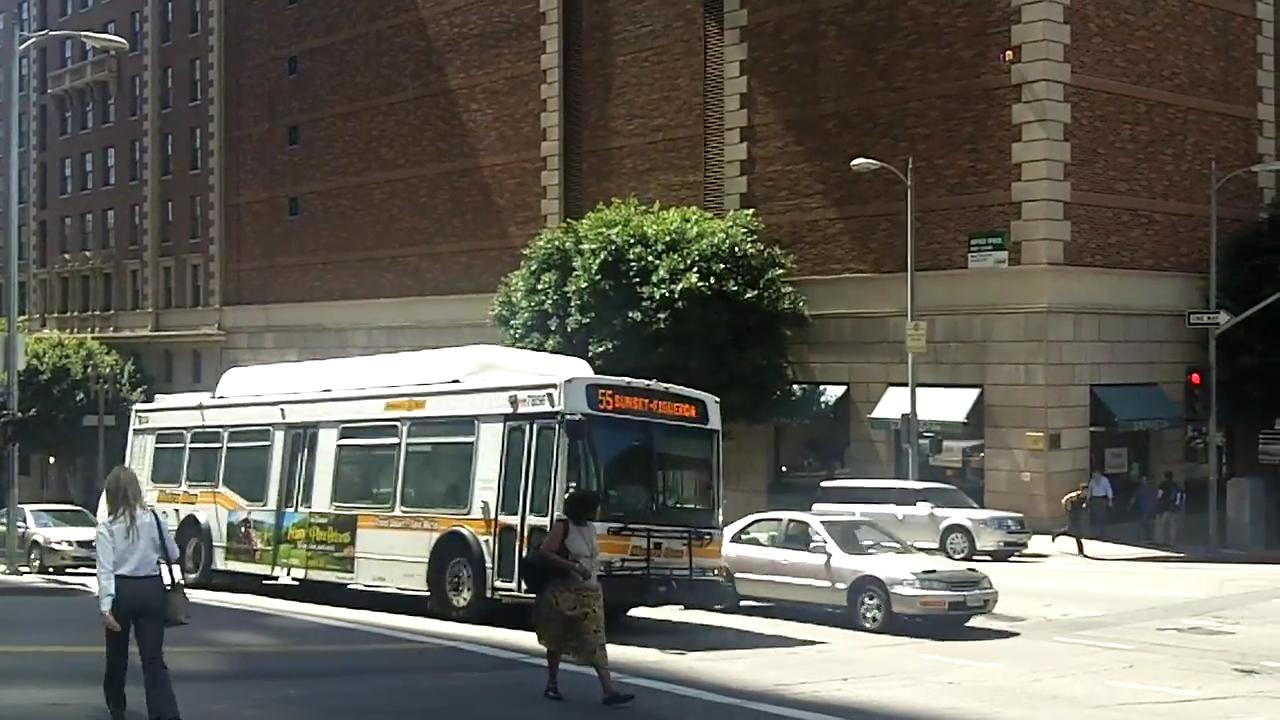}
  \includegraphics[width=.32\columnwidth]{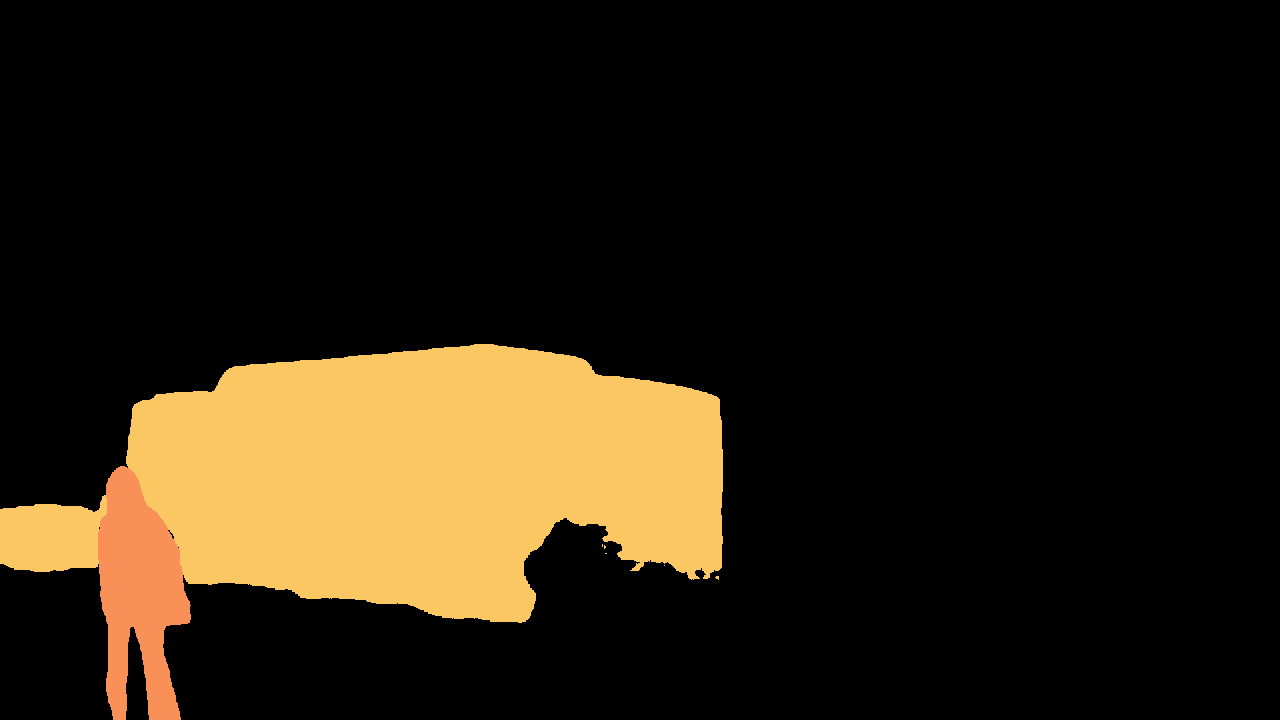}
  \includegraphics[width=.32\columnwidth]{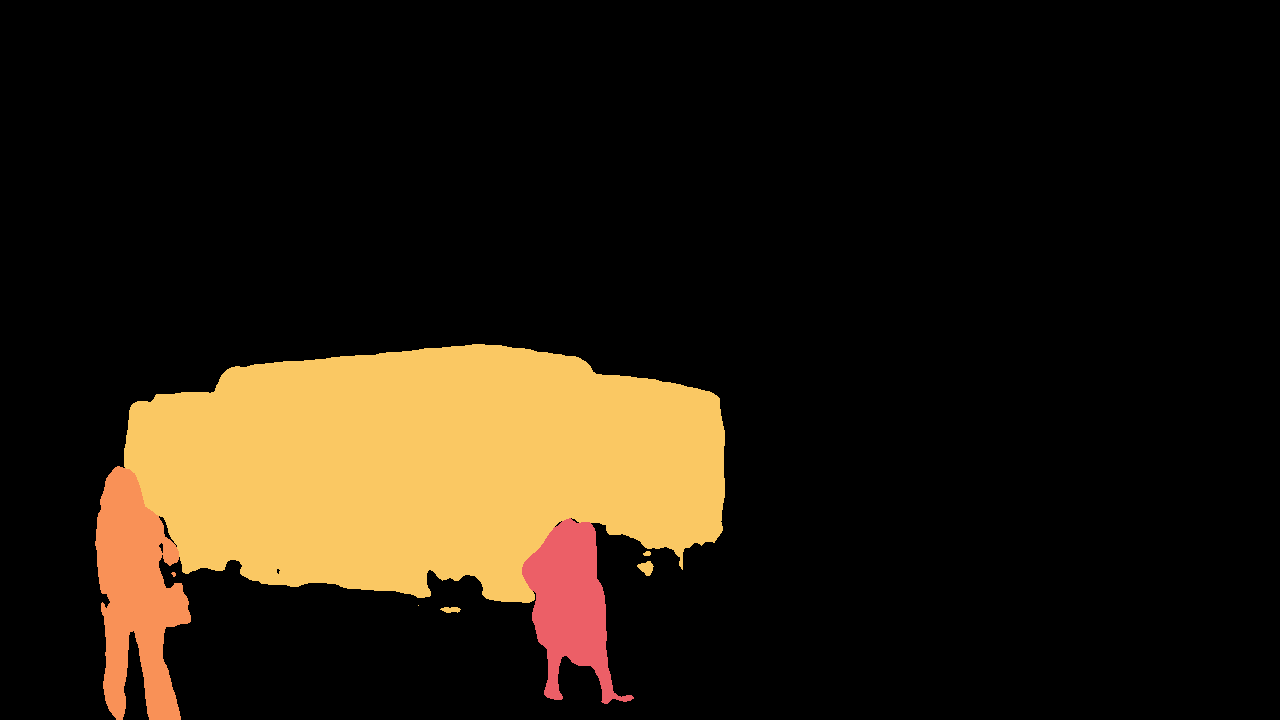}

  \includegraphics[width=.32\columnwidth]{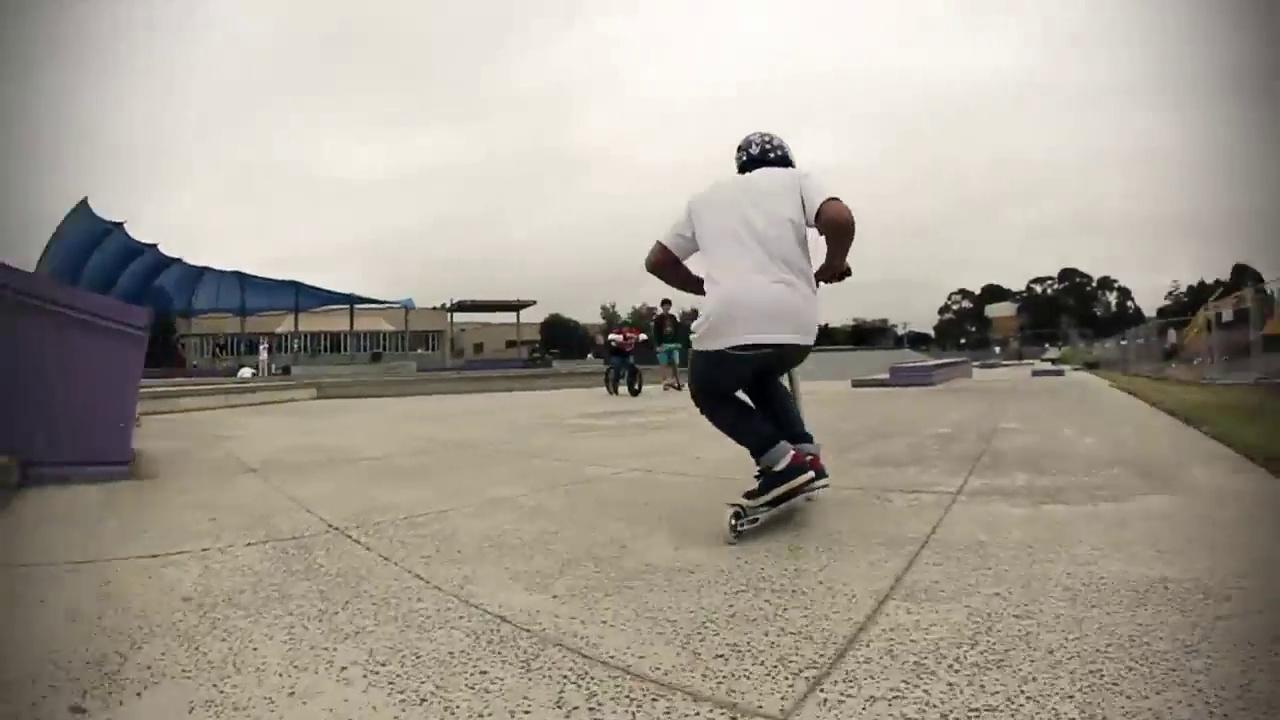}
  \includegraphics[width=.32\columnwidth]{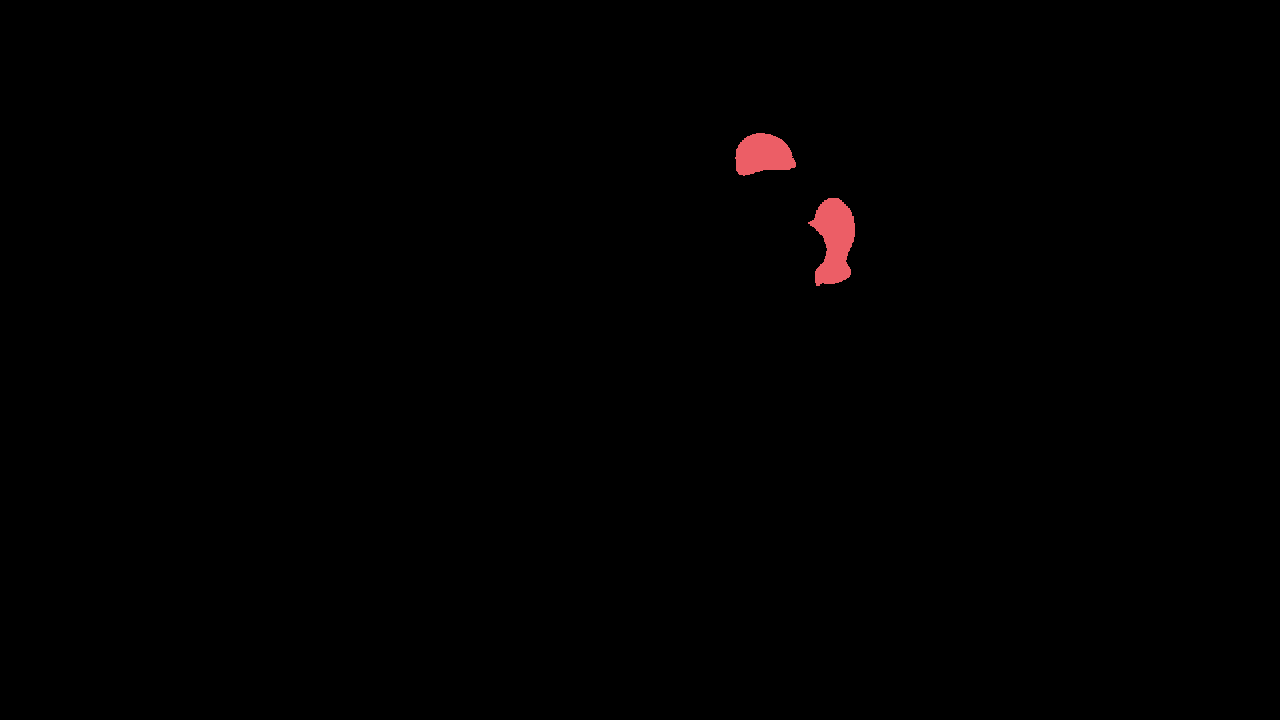}
  \includegraphics[width=.32\columnwidth]{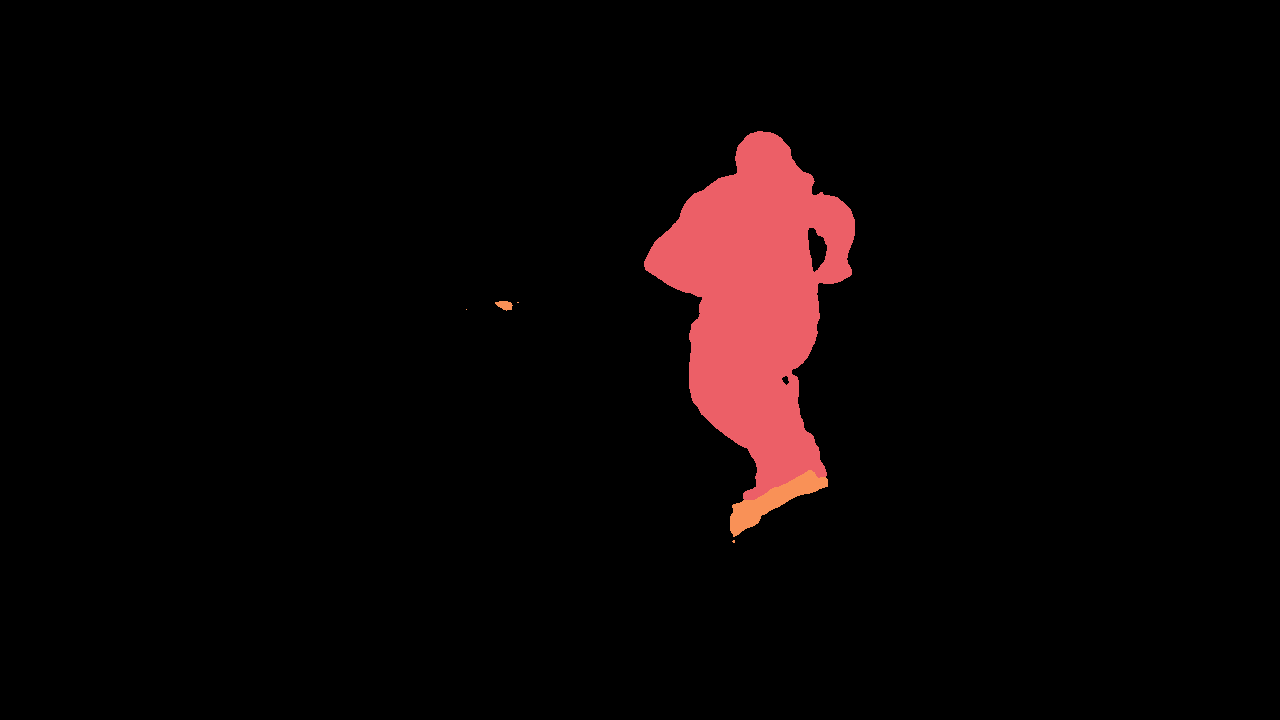}
  
  \includegraphics[width=.32\columnwidth]{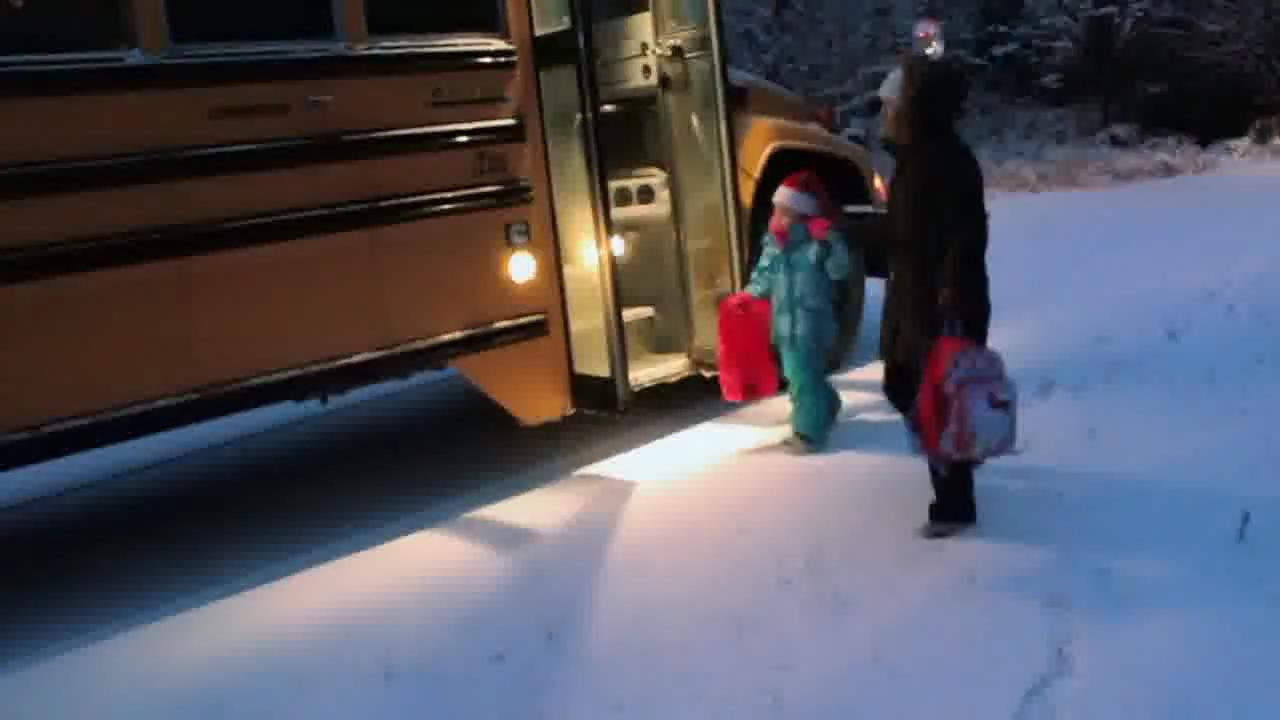}
  \includegraphics[width=.32\columnwidth]{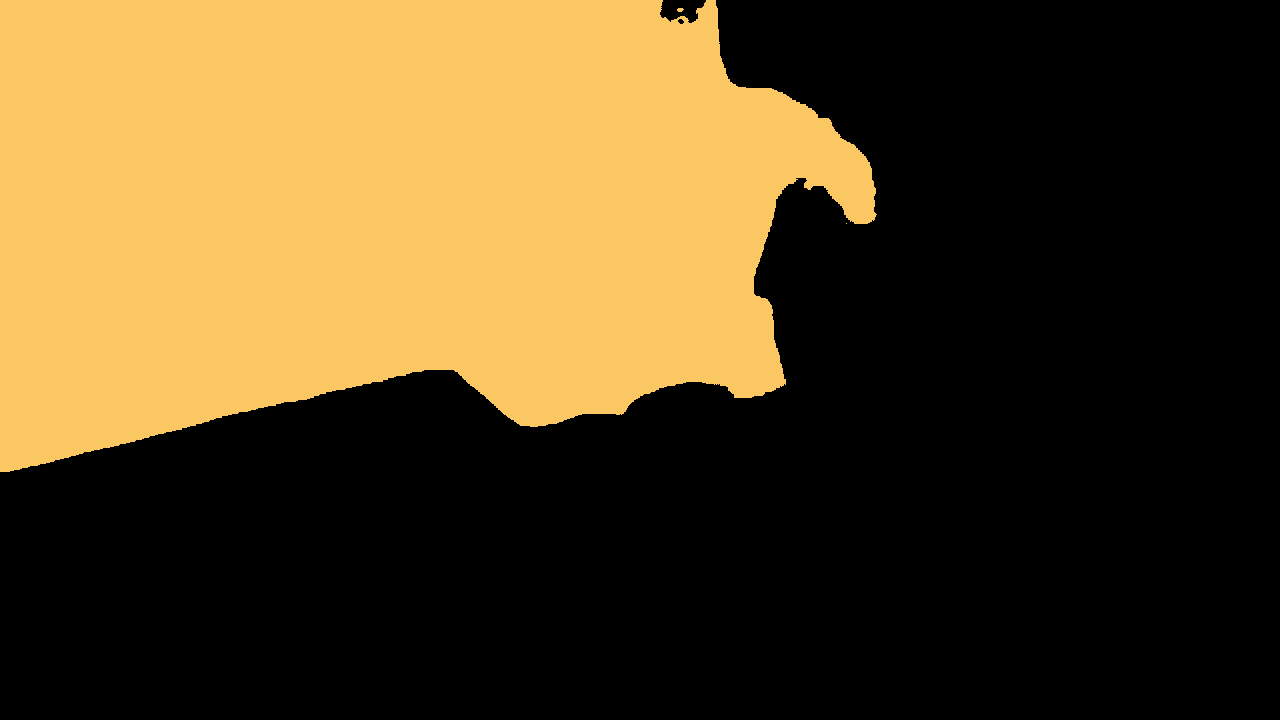}
  \includegraphics[width=.32\columnwidth]{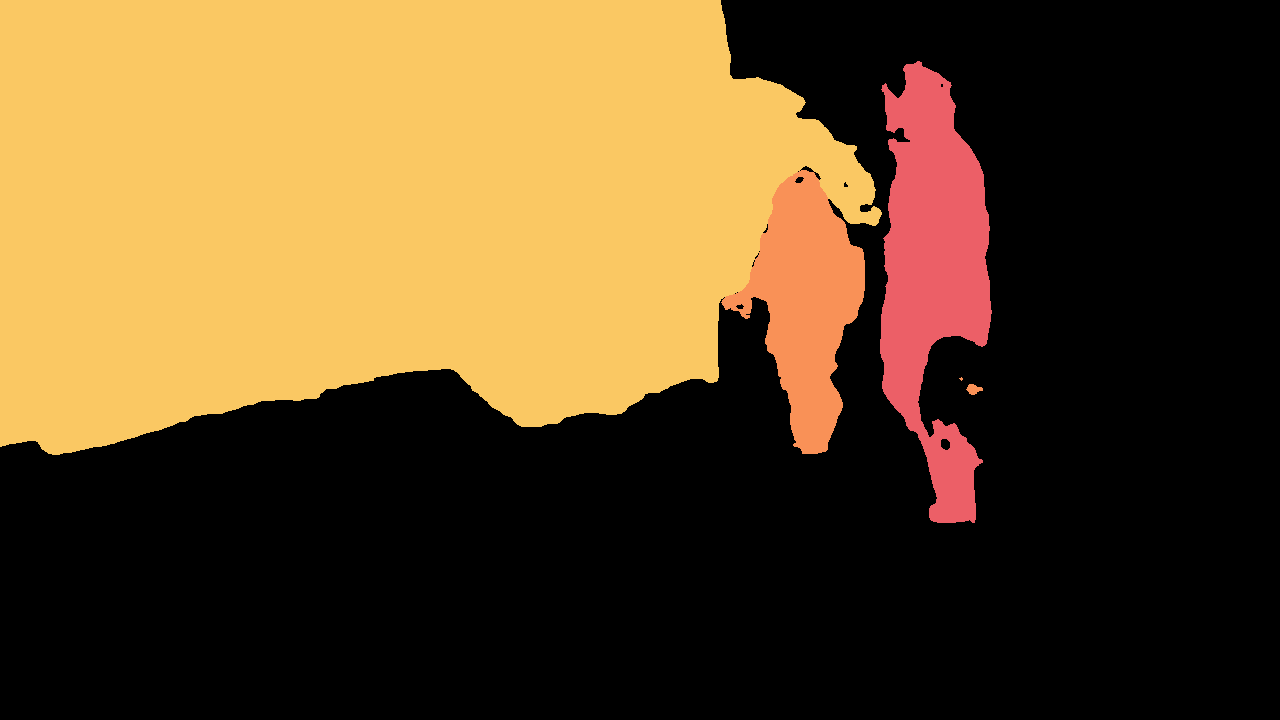}
  
  \includegraphics[width=.32\columnwidth]{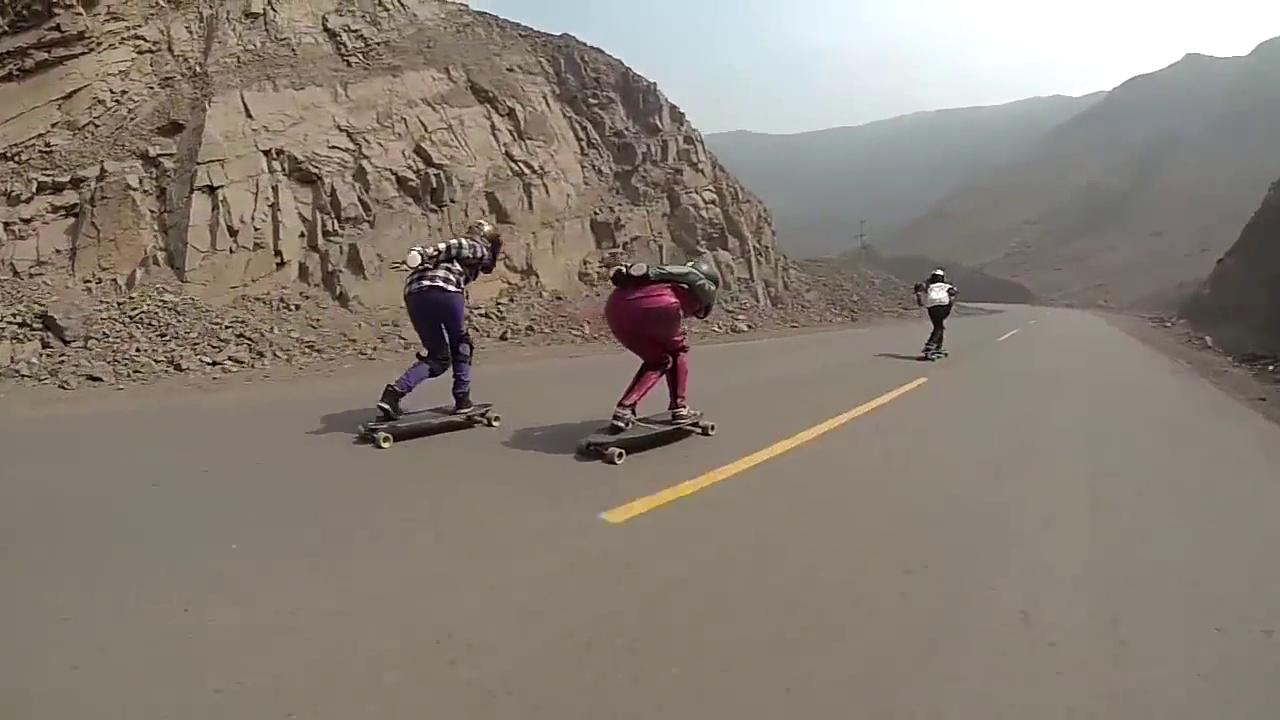}
  \includegraphics[width=.32\columnwidth]{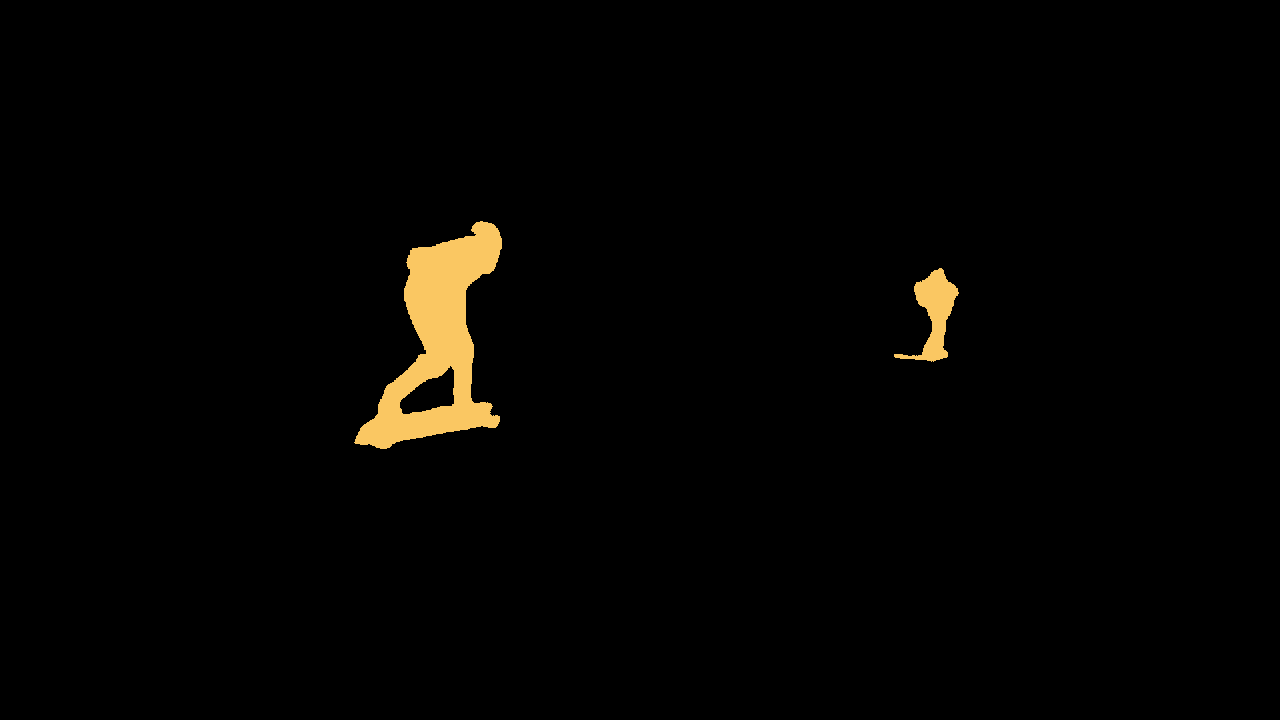}
  \includegraphics[width=.32\columnwidth]{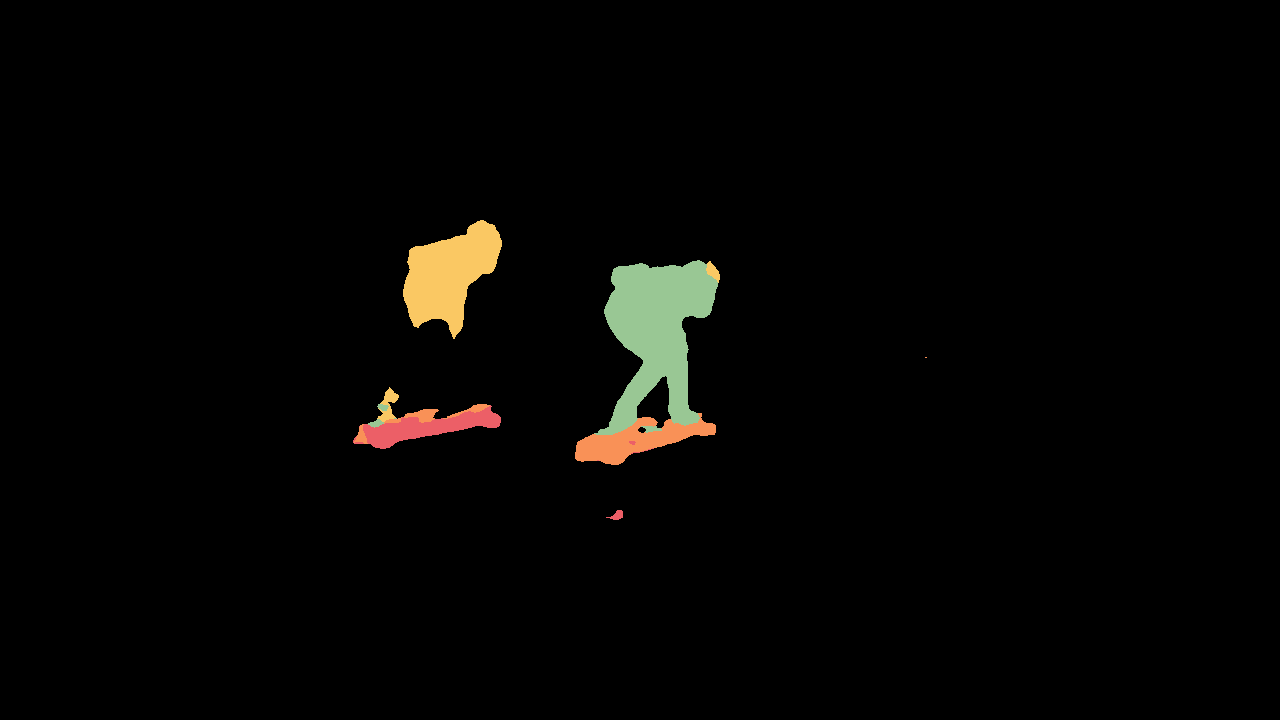}

  \parbox{.32\columnwidth}{\centering Image}
  \parbox{.32\columnwidth}{\centering RGMP \cite{RGMP}}  
  \parbox{.32\columnwidth}{\centering Ours}

  \caption{Comparison between our proposed approach and the recently proposed RGMP \protect{\cite{RGMP}}. In RGMP, the input features are concatenated with the initial mask and feature map. In contrast, we explicitly capture the target and background appearance, including distractor objects, by generative modelling. While RGMP severely struggles, the proposed approach successfully identifies and accurately segments all annotated targets. As in RGMP, we do not invoke computationally intensive fine-tuning in the first frame, but instead aim to learn the appearance model in a single forward pass. The figure is best viewed in colour.}
  \label{fig:intro}
\end{figure}

To tackle the aforementioned challenges, the standard strategy is to invoke extensive iterative optimization in the first frame \cite{OSVOS,OnAVOS,OSVOS-S,CINM}, given the initial image-mask pair. However, this strategy comes at an immense computational cost, rendering real-time operation infeasible. Furthermore, these methods do not train the segmentation pipeline end-to-end, since the online fine-tuning step is excluded from the offline learning stage. In response to the these issues, we explore the problem of finding a feedforward network architecture for video object segmentation that completely avoids online optimization. 

Recent works have posed video object segmentation as a feedforward mask-refinement process \cite{MSK,OSMN,RGMP}, where the previous mask prediction is adapted to fit the target in the current frame using a convolutional neural network. However, since no explicit modelling of the target appearance is performed, such approaches inherently fail if the target is occluded or out of view. This problem has been approached by incorporating simple appearance models based on e.g., concatenation of the feature map from the first frame \cite{RGMP}, or utilization of a set of foreground and background feature vectors \cite{VM,PML}. However, these appearance models are either too simplistic, achieving unsatisfactory discriminative power, or cannot be fully trained end-to-end due to the reliance of non-differentiable components. 

In this work, we propose a novel neural network architecture for video object segmentation that integrates a powerful appearance model of the scene. In contrast to previous methods, our network internally learns a generative probabilistic model of the foreground and background feature distributions. For this purpose, we employ a class-conditional mixture of Gaussians, which is inferred through a single forward pass. Our appearance model outputs the posterior class probabilities, thus providing a powerful cue containing discriminative information about the image content. This completely removes the need for online fine-tuning, as target-specific appearance information is captured in a single forward pass. We demonstrate our approach in fig.~\ref{fig:intro}.

The proposed generative appearance model is seamlessly integrated as a module in our video object segmentation network. Our complete architecture is composed of a backbone feature extractor, the generative appearance module, a mask propagation branch, a fusion component, and a final upsampling and prediction module. For our generative appearance module, both the model inference and the prediction stages are fully differentiable. This ensures that the entire segmentation pipeline can be trained end-to-end, which is not the case for methods invoking online fine-tuning \cite{OSVOS,OSVOS-S,OnAVOS,CINM,MSK,MGCRN} or K-Nearest-Neighbor prediction \cite{VM,PML}. Finally, our appearance module is lightweight, enabling efficient online inference. 

We perform extensive experiments on 3 datasets, including the recent large-scale YouTubeVOS dataset \cite{YTVOS}. We obtain a final score of $66.0\%$ on YouTube-VOS, outperforming all previously published methods. Further, our approach achieves the best mean IoU of $67.2\%$ on Davis17 among all causal video object segmentation methods. We perform a comprehensive analysis of our method in terms of an ablation study. Our analysis clearly underlines the effectiveness of the proposed generative appearance module and the importance of full end-to-end learning.

%% file: relatedwork.tex
\section{Related Work}
In this work we address the problem of video object segmentation where an initial segmentation mask is provided, defining the target in the first frame. In recent years interest in this problem has surged and a wide variety of approaches have been proposed. Caelles et al.\ \cite{OSVOS} proposed to use a convolutional neural network pre-trained for the semantic segmentation task, and fine-tune this in the first frame to segment out the foreground and background. This approach was extended in a number of works: continuous training during the sequence \cite{OnAVOS}; adding instance-level semantic information \cite{OSVOS-S}; incorporating motion information via optical flow \cite{MGCRN,CINM,SFL}; performing temporal propagation via a Markov random field \cite{CINM}; location-specific embeddings \cite{LSE}; sophistic data augmentation \cite{LuT}; or a combination of these \cite{PReMVOS}. While these approaches obtain satisfactory results in many scenarios, they have one critical drawback in common: they learn the target appearance in the initial frame via extensive training of deep neural networks with stochastic gradient descent. This leads to a significant time-delay before these methods can start tracking, and an average computation time that renders real-time processing infeasible.

Despite reduced accuracy, several approaches avoid invoking expensive fine-tuning procedures in the first frame.  Some methods rely on optical flow coupled with refinement \cite{CTN,OFL}. Li et al. proposed DyeNet \cite{DyeNet}, which combines optical flow with an object proposal network, interleaving bidirectional mask-propagation and target re-identification. DyeNet provides outstanding performance, but it is not causal and relies on future video frames to make predictions. Jampani et al.\ \cite{VPN} explicitly try to avoid optical flow and propose an approach based on bilateral filters. Cheng et al.\ \cite{FAVOS} track different parts of the target with visual object tracking techniques, and refine the final solution with a convolutional neural network. Xu et al.\ \cite{YTVOS} instead train a convolutional LSTM \cite{hochreiter1997lstm} to track and segment the target.

More closely related to our work, Perazzi et al.\ \cite{MSK} pose video object segmentation as a mask refinement problem. Based on an input image, the mask predicted from the previous frame is refined with a neural network. The network is recurrent in time, with a particularly deep recurrent connection, an entire VGG16 \cite{SimonyanICLR2015}. In the work by Yang et al. \cite{OSMN}, the mask was reduced to a rough spatial prior on the target location, and this together with a channel-wise attention mechanism provided improved performance. Wug et al. \cite{RGMP} extend \cite{MSK} and concatenate the initial frame feature map and mask with the current feature map and previous mask, and train a standard convolutional neural network to match and segment in a fully recurrent fashion. Also more explicit matching mechanisms have been proposed, where the input features are matched with a set of features with known class membership \cite{VM,PML} using K-Nearest-Neighbour (KNN). While these methods model the target appearance, the non-parametric nature of KNN requires the entire training set to be stored. Additionally, the process of finding the K nearest neighbours is not differentiable. In contrast to existing work, our approach learns a compact appearance model of the scene in a single differentiable forward pass.

%% file: method.tex
\section{Method}\label{sec:method}
\begin{figure*}[ht]
  \includegraphics[width=\textwidth]{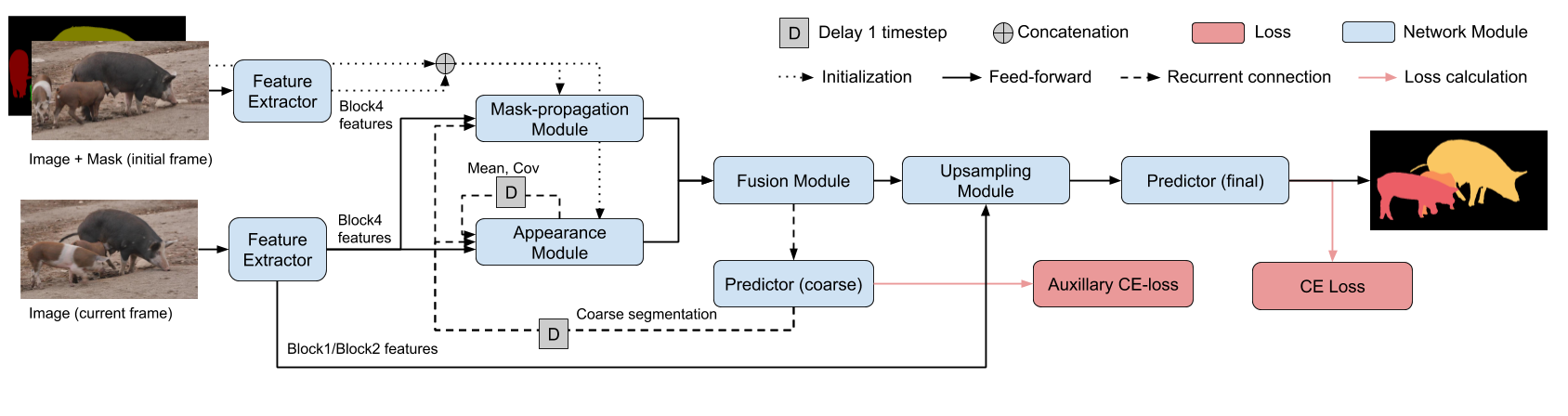}
  \caption{Full architecture of the proposed approach, illustrating both model initialization and frame processing. Model Initialization: A feature map is extracted from the initial frame, which is then fed together with the mask to the mask propagation module. This pair is furthermore used to initialize the appearance model. Frame processing: A feature map is extracted from the current frame and fed to both the appearance and mask-propagation modules whose outputs are combined, generating a coarse mask-encoding. Our upsampling module then refines the mask-encoding by also considering low-level information contained in the shallow features. The predictor then generates a final segmentation, based on this encoding. Moreover, the mask-encoding and appearance model parameters are fed back via a recurrent connection. During training, we use two cross-entropy losses applied to the coarse and fine segmentations, respectively.}

  \label{fig:architecture}
\end{figure*}

The aim of this work is to develop a network architecture for video object segmentation with the capability of learning accurate models of the target and background appearance through a single forward pass. That is, the network must learn in a one-shot manner to discriminate between target and background pixels, without invoking stochastic gradient descent. We tackle this problem by integrating a generative model of the foreground and background appearance. This model directly aids the segmentation process by providing discriminative posterior class probabilities. The learning and inference is computationally efficient and end-to-end differentiable, enabling a seamless integration of our generative component into a neural network.

\subsection{Overview}
Our approach is divided into five components that jointly address the video object segmentation task and are trained jointly end-to-end. The model is illustrated in fig.~\ref{fig:architecture}. Given an input image, features are first extracted with a backbone network. These are then passed to the appearance- and mask-propagation modules. The outputs of these two modules are combined in the fusion module, comprising two convolutional layers and outputting a coarse mask encoding. The encoding is handed to a predictor that generates a coarse segmentation mask. This prediction is used to update the appearance module and further used as input to the mask-propagation layer in the next frame to provide a rough spatial prior. The mask encoding output by the fusion component is also passed through an upsampling module, in which the coarse encoding is combined with successively more shallow features in order to produce a final refined segmentation.

\subsection{Generative Appearance Module}
\label{sec:appearance-module}
The task of our appearance module is to learn a generative model of the video content in a deep feature space. Our generative model is \emph{conditioned} on the class variable, indicating target or background. Given a new frame, the appearance module returns the posterior class probabilities at each image location. This output forms an extremely strong cue for foreground/background discrimination, as the proposed module explicitly models their respective appearance in a probabilistic manner.

\parsection{Model learning}
Formally, let the set of features extracted from the image be denoted as $\{\vecn{x}_p\}_p$. The feature $\vecn{x}_p$ at each spatial location $p$ is a $D$-dimensional vector of real numbers. We model these observed feature vectors as i.i.d.\ samples drawn from the underlying distribution
\begin{equation}
  p(\vecn{x}_p) = \sum_{k=1}^K p(z_p=k)p(\vecn{x}_p|z_p=k) \enspace.
  \label{marginal}
\end{equation}
Each class-conditional density is a multi-variate Gaussian with mean $\bm{\mu}_k$ and covariance matrix $\bm{\Sigma}_k$, 
\begin{equation}
\label{eq:gauss-cond}
  p(\vecn{x}_p|z_p=k) = \mathcal{N}(\vecn{x}_p|\bm{\mu}_k,\bm{\Sigma}_k) \enspace.
\end{equation}
The discrete random variable $z_p$ in \eqref{marginal} assigns the observation $\vecn{x}_p$ to a specific component $z_p = k$. We use a uniform prior $p(z_p=k) = 1/K$ for this variable, where $K$ is the number of components. In our model, we use separate components $z_p = k$ to describe the foreground and background respectively. Each $z_p = k$ is thus strictly assigned to either foreground or background, making our model conditioned on the class. As further detailed below, we use two Gaussian components for each class.

In the first frame, our generative mixture model is inferred from the extracted features and the initial target mask. In subsequent frames, we update the model using the network predictions as soft class labels. In general, to update the mixture model in a frame $i$ we require a set of features $\vecn{x}_p^i$ together with a set of soft component assignment variables $\alpha_{pk}^i\in[0,1]$. These variables can be thought of as soft labels, describing the level of assignment of the vector $\vecn{x}_p^i$ to component $k$. For example, in the first frame $i=0$, the feature vectors would be strictly assigned to either foreground or background $\alpha_{pk}^0\in\{0,1\}$ using the initial target mask. 

Given the variables $\alpha_{pk}^i$, we compute the model parameter updates as,
\begin{subequations}
	\label{eq:gen-param}
	\begin{align}
		\bm{\tilde{\mu}}_k^{i} &= \frac{\sum_p \alpha_{pk}^i\vecn{x}_p^i}{\sum_p \alpha_{pk}^i}\enspace,\label{muest}\\
		\bm{\tilde{\Sigma}}_k^{i} &= \frac{\sum_p \alpha_{pk}^i\diag\{(\vecn{x}_p^i - \bm{\tilde{\mu}}_k^{i})^2 + \vecn{r}_k\}}{\sum_p \alpha_{pk}^i}\enspace.\label{covest}
	\end{align}
\end{subequations}
For efficiency, we limit the covariance matrix to be diagonal, where $\mathrm{diag}(\vecn{v})$ is a diagonal matrix with entries corresponding to the input vector $\vecn{v}$. To avoid singularities, the covariance is regularized with a vector $\vecn{r}_k$, which is a trainable parameter in our network. In the first frame, the mixture model parameters in \eqref{eq:gauss-cond} are directly achieved from \eqref{eq:gen-param}, i.e.\ $\bm{\mu}_k^{0} = \bm{\tilde{\mu}}_k^{0}$ and $\bm{\Sigma}_k^{0} = \bm{\tilde{\Sigma}}_k^{0}$. In subsequent frames, these parameters are updated with new information \eqref{eq:gen-param} using a learning rate $\lambda$,
\begin{align}
\label{eq:param-update}
\bm{\mu}_k^i &= (1-\lambda)\bm{\mu}_k^{i-1} + \lambda\bm{\tilde{\mu}}_k^i\enspace, \nonumber\\
\bm{\Sigma}_k^i &= (1-\lambda)\bm{\Sigma}_k^{i-1} + \lambda\bm{\tilde{\Sigma}}_k^i\enspace.
\end{align}

\parsection{Assignment variables}
Next, we describe the computation of the assignment variables $\alpha_{pk}^i$. Note that \eqref{eq:gen-param} resembles the M-step in the Expectation Maximization (EM) algorithm for a mixture of Gaussians. In EM, the variables $z_p^i$ are treated as latent and \eqref{eq:gen-param} is derived by maximizing the expected complete-data log-likelihood. In that case the assignment variables are computed in the E-step as  $\alpha_{pk}^i = p(z_p^i = k | \vecn{x}_p^i, \theta^{i-1})$, where $\theta^{i-1} = \{\bm{\mu}_k^{i-1}, \bm{\Sigma}_k^{i-1}\}_k$ are the previous estimates of the parameters. However, the setting is different in our case. The discrete assignment variables $z_p^i$ are fully observed in the first frame. Moreover, in the subsequent frames, the network refines the posteriors $p(z_p^i = k | \vecn{x}_p^i, \theta^{i-1})$, providing even better assignment estimates. We therefore exploit these factors in the estimation of the assignment variables $\alpha_{kp}^i$.

Our model consists of one \emph{base component} for background $k=0$ and foreground $k=1$, respectively. Given the ground truth binary target mask in the first frame $y_p$, where $y_p=1$ for foreground and $y_p=0$ otherwise, we set $\alpha_{p0}^0 = 1 - y_p$ and $\alpha_{p1}^0 = y_p$. That is, the feature vectors $x_p^i$ are strictly assigned to the foreground and background base components according to the initial mask. 
In subsequent frames, where the ground-truth is not available, we use the final prediction of our segmentation network according to
\begin{align}
\label{eq:aff-pred}
\alpha_{p0}^i &= 1 - \tilde{y}_p(I^i,\theta^{i-1},\Phi)\nonumber\\
\alpha_{p1}^i &= \tilde{y}_p(I^i,\theta^{i-1},\Phi) \enspace.
\end{align}
Here, $\tilde{y}_p(I^i,\theta^{i-1},\Phi)\in[0,1]$ is the probability of the target class, given the input image $I^i$, neural network parameters $\Phi$, and current mixture model parameter estimates $\theta^{i-1}$.

A drawback of using a single Gaussian component per class is that only uni-modal distributions can be accurately represented. However, the background appearance is typically multi-modal, especially in the presence of background objects that are similar to the target, often termed \emph{distractors}. To obtain satisfactory discrimination between foreground and background, it is therefore critical to capture the feature distribution of such distractors. We therefore add Gaussian components in our model that are dedicated to the task of modeling hard examples. These components are explicitly learned to counter the errors of the two base components. Ideally, we would wish the base components alone to correctly predict the assignment variables, i.e.\ $p(z_p^i = k|\vecn{x}_p^i,\bm{\mu}_k^i,\bm{\Sigma}_k^i) = \alpha_{pk}^i,\, k=0,1$. The additional components are trained on data where this does not hold by considering incorrectly classified background ($k=2$) and foreground ($k=3$) respectively. Their corresponding assignment variables are computed as,
\begin{align}
\label{eq:residual}
\alpha_{p2}^i &= \relu(p(z_p^i = 0|\vecn{x}_p^i,\bm{\mu}_0^i,\bm{\Sigma}_0^i) - \alpha_{p0}^i)\nonumber\\
\alpha_{p3}^i &= \relu(p(z_p^i = 1|\vecn{x}_p^i,\bm{\mu}_1^i,\bm{\Sigma}_1^i) - \alpha_{p1}^i)\enspace.
\end{align}
Here, the posteriors $p(z_p^i = k|\vecn{x}_p^i,\bm{\mu}_k^i,\bm{\Sigma}_k^i)$ are evaluated using only the base components. Given \eqref{eq:residual}, we finally update the parameters of the latter components $k=2,3$ using \eqref{eq:gen-param} and \eqref{eq:param-update}.

\parsection{Module output}
Given the mixture model parameters computed in the previous frame, $\theta^{i-1}$, our model can predict the component posteriors,
\begin{equation}
\label{eq:posteriors}
p(z_p^i = k|\vecn{x}_p^i, \theta^{i-1}) = \frac{p(z_p^i=k)p(\vecn{x}_p^i|z_p^i = k)}{\sum_k p(z_p^i=k)p(\vecn{x}_p^i|z_p^i = k)}\enspace.
\end{equation}
Note that each component $k$ belongs to either foreground or background, and that the outputs \eqref{eq:posteriors} thus provide a discriminative mask encoding. In practice, we found it beneficial to feed the log-probabilities $\log(p(z_p^i=k)p(\vecn{x}_p^i|z_p^i = k))$ into the conv layers in the fusion module. By canceling out constant factors, the outputs are calculated as, 
\begin{equation}
\label{eq:scores}
s_{pk}^i = -\frac{\ln |\bm{\Sigma}_k^{i-1}| + (\vecn{x}_p^i - \bm{\mu}_k^{i-1})\tp(\bm{\Sigma}_k^{i-1})^{-1}(\vecn{x}_p^i - \bm{\mu}_k^{i-1})}{2}.
\end{equation}
The component posteriors \eqref{eq:posteriors} can be reconstructed from $s_{pk}^i$ by a simple soft-max operation. The output \eqref{eq:scores} should therefore be interpreted as component \emph{scores}, encoding foreground and background assignment. The entire appearance modelling procedure is summarized in Algorithm~\ref{alg:appearance-module}.

\subsection{Object Segmentation Architecture}
\label{sec:architecture}
As our backbone feature extractor, we use ResNet101 \cite{he2016deep} with dilated convolutions \cite{chen2018deeplab} to reduce the stride of the deepest layer from 32 to 16. It is pretrained on ImageNet and all layers up to the last block, \texttt{layer4}, are frozen. The \textit{mask-propagation module} is based on the concept proposed in \cite{RGMP}. The module constructs a mask encoding based on the mask predicted in the previous frame, a feature map predicted in the current frame, and a feature map extracted from the initial frame together with the given ground-truth mask. The entire module consists of three convolutional layers, where the middle layer is a dilation pyramid \cite{chen2018deeplab}.

The outputs of the mask propagation and appearance modules are concatenated and fed into the \emph{fusion module}, comprising two convolutional layers. The output is then sent as input to the \emph{upsampling module} from which a predicted soft segmentation $\hat{y}_p$ is obtained. The output of the fusion module is also fed into a predictor that produces a coarse segmentation $\tilde{y}_p$, which is input to the mask propagation step and appearance module (using \eqref{eq:aff-pred}) in the next timestep. By separating the feature extractor and upsampling path from the recurrent module we get a shorter path between variables of different time steps. We experienced the coarse mask to be a sufficient representation of the previous target segmentation. As a special case, during sequences with multiple objects, we run our approach once per object, and combine the resulting soft segmentations with softmax-aggregation \cite{RGMP}. The aggregated soft segmentations then replaces the coarse segmentation $\tilde{y}_p$ in the recurrent connection.

The output of the fusion module provides coarse mask-encoding that is used to locate and segment the target. There have been considerable efforts in semantic segmentation and instance segmentation litterature to refine final segmentations. We adopt an upsampling path similar to \cite{pinheiro2016learning}, where the coarse representation is successively combined with successively shallower features.

\begin{algorithm}[!t]
  \caption{The appearance module inference and update. Inference: Based on the appearance model parameters, $\bm{\mu}_k^i, \bm{\Sigma}_k^i$, and the input feature map $\vecn{x}_p^i$, a soft segmentation is constructed for the background, foreground, and the two residual components. Update: The appearance model parameters are updated based on the coarse segmentation $\tilde{y}_p^i$.}

  \label{alg:appearance-module}
  \DontPrintSemicolon
  \SetKwProg{Fn}{}{:}{}
  \SetKwFunction{FInference}{Inference}
  \Fn{\FInference{$\vecn{x}_p^i$, $\bm{\mu}_k^i$, $\bm{\Sigma}_k^i$}}{
    for $k=0,1,2,3$: compute $s_{pk}^i$ from \eqref{eq:scores}\;
    \KwRet $s_{pk}^i$\;
  }

  \SetKwFunction{FUpdate}{Update}  
  \Fn{\FUpdate{$\vecn{x}_p^i$, $\tilde{y}_p^i$, $\bm{\mu}_k^i$, $\bm{\Sigma}_k^i$}}{
  for $k=0,1$: compute $\alpha_{pk}^{i}$ from \eqref{eq:aff-pred}\;
  for $k=0,1$: compute $\tilde{\bm{\mu}}_k^{i}, \tilde{\bm{\Sigma}}_k^{i}$ based on \eqref{eq:gen-param}\;
  for $k=0,1$: compute $s_{pk}^{i}$ based on \eqref{eq:scores}\;
  for $k=0,1$: compute $p(z_p^i = k|\vecn{x}_p^i,\bm{\mu}_0^i,\bm{\Sigma}_0^i) = \softmax(s_{p0}^{i},s_{p1}^{i})$\;

  for $k=2,3$: compute $\alpha_{pk}^{i}$ from \eqref{eq:residual}\;
  for $k=2,3$: compute $\tilde{\bm{\mu}}_k^{i}, \tilde{\bm{\Sigma}}_k^{i}$ based on \eqref{eq:gen-param}\;
  for $k=0,1,2,3$: update $\bm{\mu}_k^{i}$ and $\bm{\Sigma}_k^{i}$ from \eqref{eq:param-update}\;
  \KwRet $\bm{\mu}_k^{i}$ and $\bm{\Sigma}_k^{i}$\;
  }
\end{algorithm}

\subsection{Network Training}
We train the proposed neural network end-to-end in a recurrent fashion. Based on a video and a single ground-truth segmentation, the network predicts segmentation masks for each frame in the video. We train on three datasets:

\parsection{DAVIS2017 \cite{DAVIS17}}
The DAVIS2017 training set comprises 60 videos containing one or several annotated objects to track. Each video is between 25 and 100 frames long, each of which is labeled with a ground-truth segmentation.

\parsection{YouTube-VOS \cite{YTVOS}}

The YouTube-VOS training set consists of 3471 videos with one or several target objects. Each video is 20 to 180 frames long, where every fifth frame is labelled. We use only the labelled frames during training.

\parsection{SynthVOS}
In order to cover a wide varity of classes we follow \cite{MSK,RGMP} and utilize objects from the salient object segmentation dataset MSRA10k \cite{cheng2015msra10k}. It contains $10^4$ images where a single object is segmented. We paste 1 to 5 such objects onto an image from VOC2012 \cite{everingham2015pascal}. A synthetic video is obtained by moving the objects across the image.

One training sample consists of a video snippet of $n$ frames and a given ground-truth for the first frame. Images are normalized with ImageNet \cite{ILSVRC15} mean and standard deviation. We let our model predict segmentation masks in each frame and apply a cross-entropy loss. We also place an auxillary loss on the coarse segmentations $\tilde{y}_p$. The losses are summed and minimized with Adam in two stages:

\parsection{Initial training}
First we train for 80 epochs using all three datasets on half resolution images ($240\times 432$). The batchsize is set to 4 video snippets, using 8 frames in each snippet. We use a learning rate of $10^{-4}$, exponential learning rate decay of $0.95$ per epoch, and a weight decay of $10^{-5}$.

\parsection{Finetuning}
We then finetune for 100 epochs on the DAVIS2017 and YouTube-VOS training sets, using full image resolution. During this step we sample sequences from both datasets with equal probability. The batchsize is lowered to 2 snippets, to accomodate longer sequences of 14 frames. We use a learning rate of $10^{-5}$, exponential learning rate decay of $0.985$ per epoch, and a weight decay of $10^{-6}$. The training is stopped early by observing the performance on a held-out set of 300 sequences from the YouTube-VOS training set.

%% file: experiments.tex
\newcommand{\yesmark}{\checkmark}
\newcommand{\nomark}{$\times$}

\newcommand{\first}[1]{#1}

\section{Experiments}

We first conduct an ablation study of the proposed approach on the Youtube-VOS benchmark \cite{YTVOS}. Then we compare with the state-of-the-art on three video object segmentation benchmarks \cite{DAVIS16,DAVIS17,YTVOS}. Our method is implemented in PyTorch \cite{pytorch} and trained on a single Nvidia V100 GPU. Our code will be made available upon publication.

\subsection{Ablation Study}
\label{sec:ablation}
We perform an extensive ablative analysis of our approach on the large-scale YouTube-VOS dataset. We use the official validation set, since the test set is closed for submissions after the challenge. The validation set comprises 474 videos, each labelled with one or multiple objects. Ground-truth masks are withheld, and results are obtained through an online evaluation server. Performance is measured in terms of the mean Jaccard index $\mathcal{J}$ \cite{DAVIS17}, i.e.\ intersection-over-union (IoU), and the mean contour accuracy $\mathcal{F}$. The two measures are separately calculated for seen and unseen classes, resulting in four performance measures. The overall performance ($\mathcal{G}$) is the average of all four measures. 

In our ablative experiments, we analyze six key modifications of our approach, as explained below. Results are shown in table~\ref{tab:ablation}. For each version, we retrain the entire network from scratch using the exact same procedure.

\begin{table}[!b]
	\centering
	\resizebox{\columnwidth}{!}{%
		\begin{tabular}{l|c|c|c}
			Version & $\mathcal{G}$ & $\mathcal{J}$ seen (\%) & $\mathcal{J}$ unseen (\%)\\\hline
			\textbf{Ours} & \first{66.0} & \first{66.9} & \first{61.2}\\
			No appearance module & 50.0 & 57.8 & 40.6\\
			No mask-prop module & 64.0 & 65.5 & 59.5\\
			Unimodal appearance & 64.4 & 65.8 & 58.8\\
			No update & 64.9 & 66.0 & 59.8\\
			Appearance SoftMax & 55.8 & 59.3 & 50.7\\
			No end-to-end & 58.8 & 62.5 & 53.1\\
		\end{tabular}
	}\vspace{1mm}

	\caption{Ablation study on YouTube-VOS. We report the overall performance $\mathcal{G}$ along with segmentation accuracy $\mathcal{J}$ on classes seen and unseen during training. See text for further details.}
	\label{tab:ablation}
\end{table}

\begin{figure}[!t]
	\parbox{.32\columnwidth}{\centering\small Image}%
	\parbox{.32\columnwidth}{\centering\small Final segmentation}%
	\parbox{.32\columnwidth}{\centering\small Appearance}
	
	\vspace{.1cm}
	
	\includegraphics[width=.32\columnwidth]{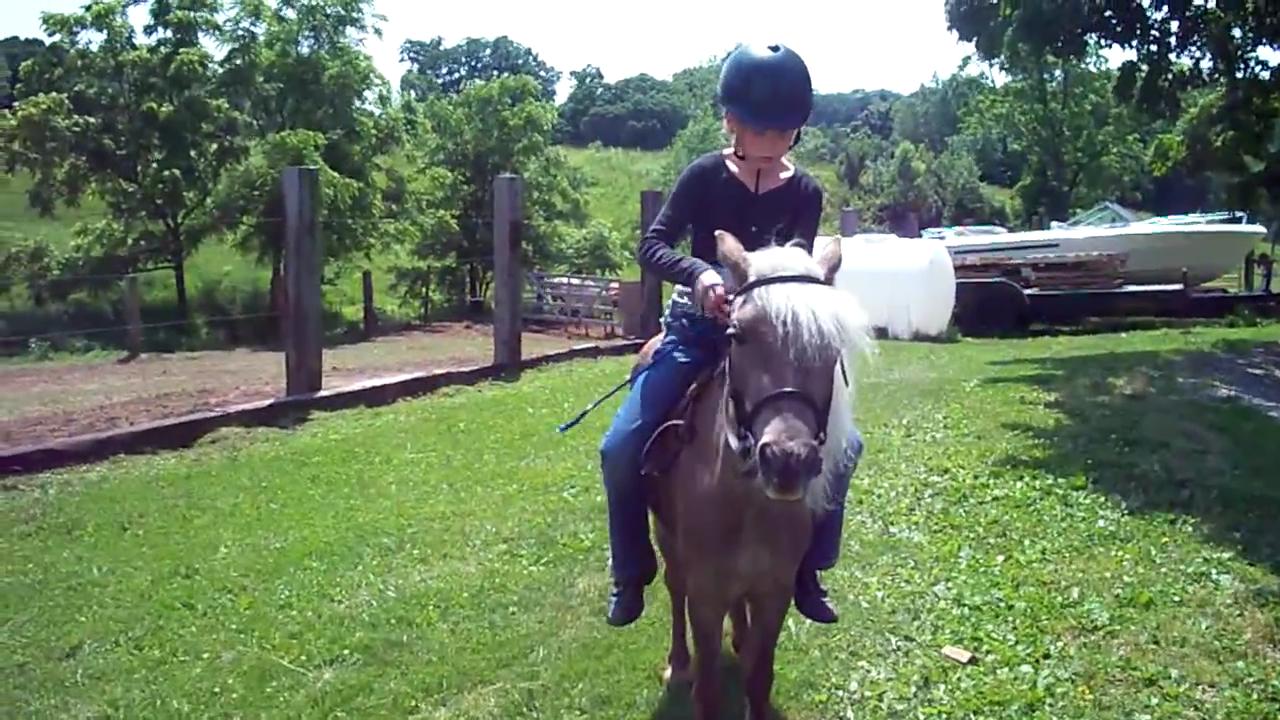}
	\includegraphics[width=.32\columnwidth]{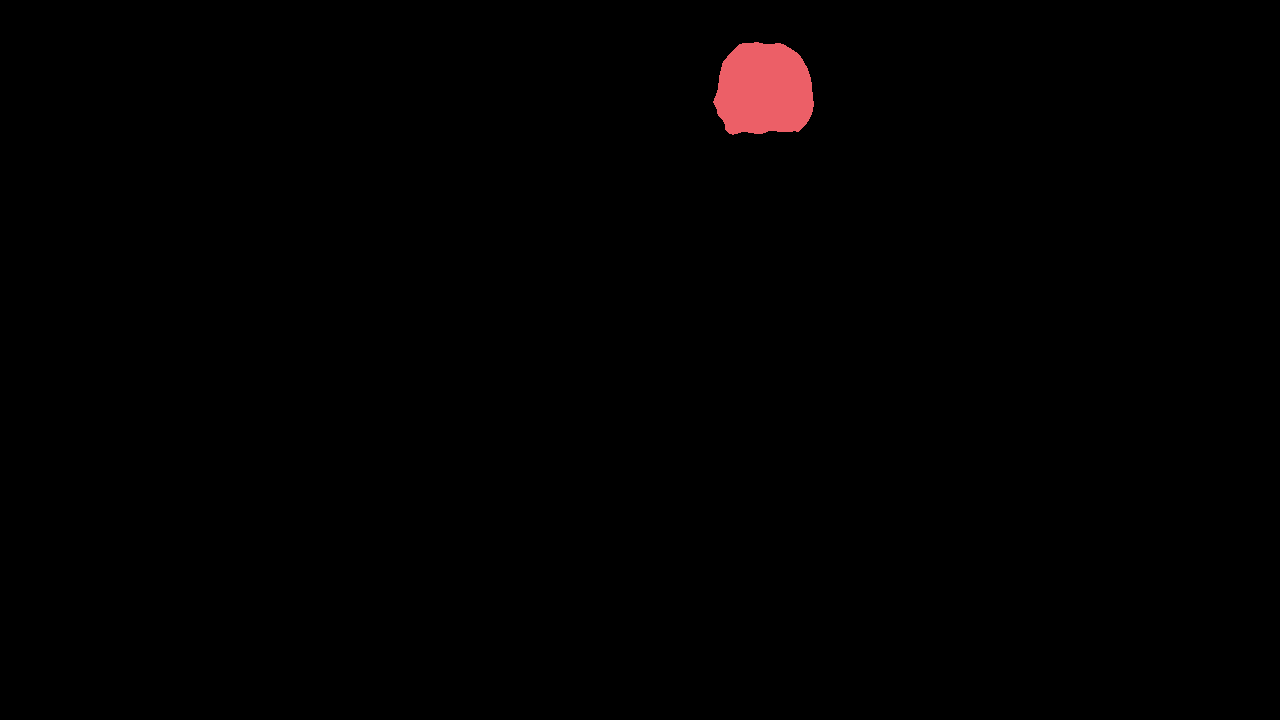}
	\includegraphics[width=.32\columnwidth]{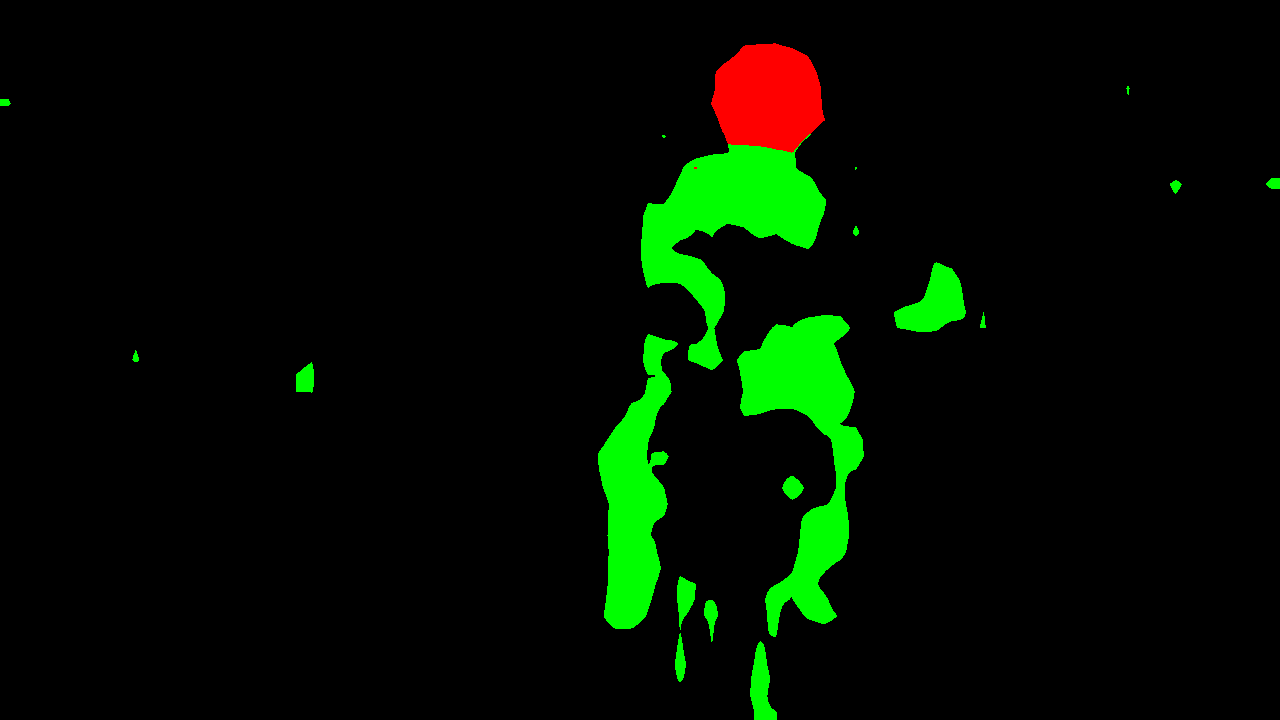}
	
	\includegraphics[width=.32\columnwidth]{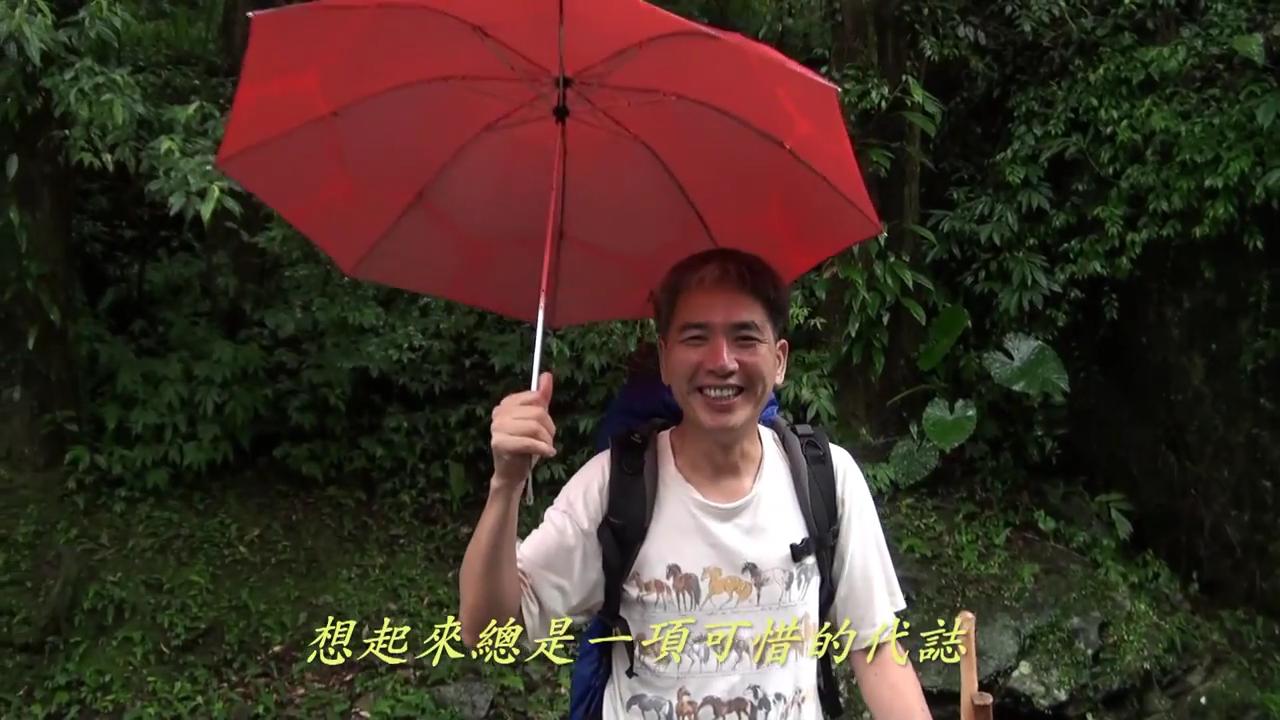}
	\includegraphics[width=.32\columnwidth]{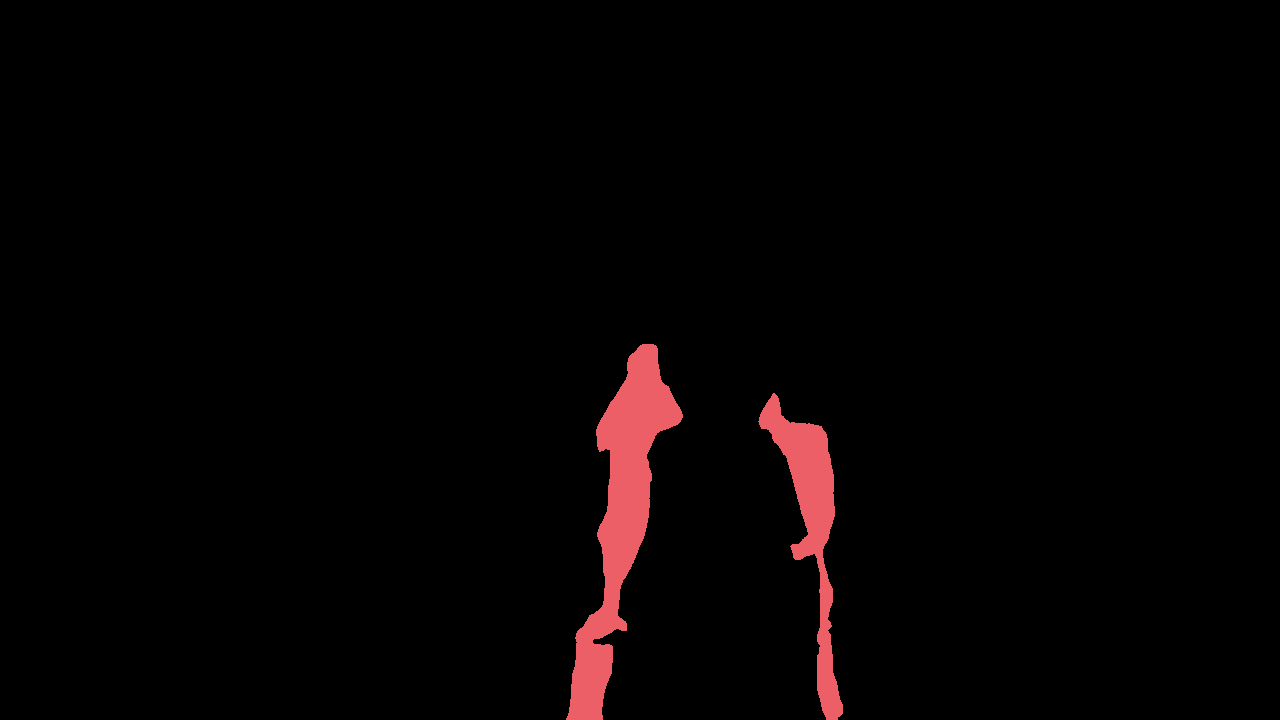}
	\includegraphics[width=.32\columnwidth]{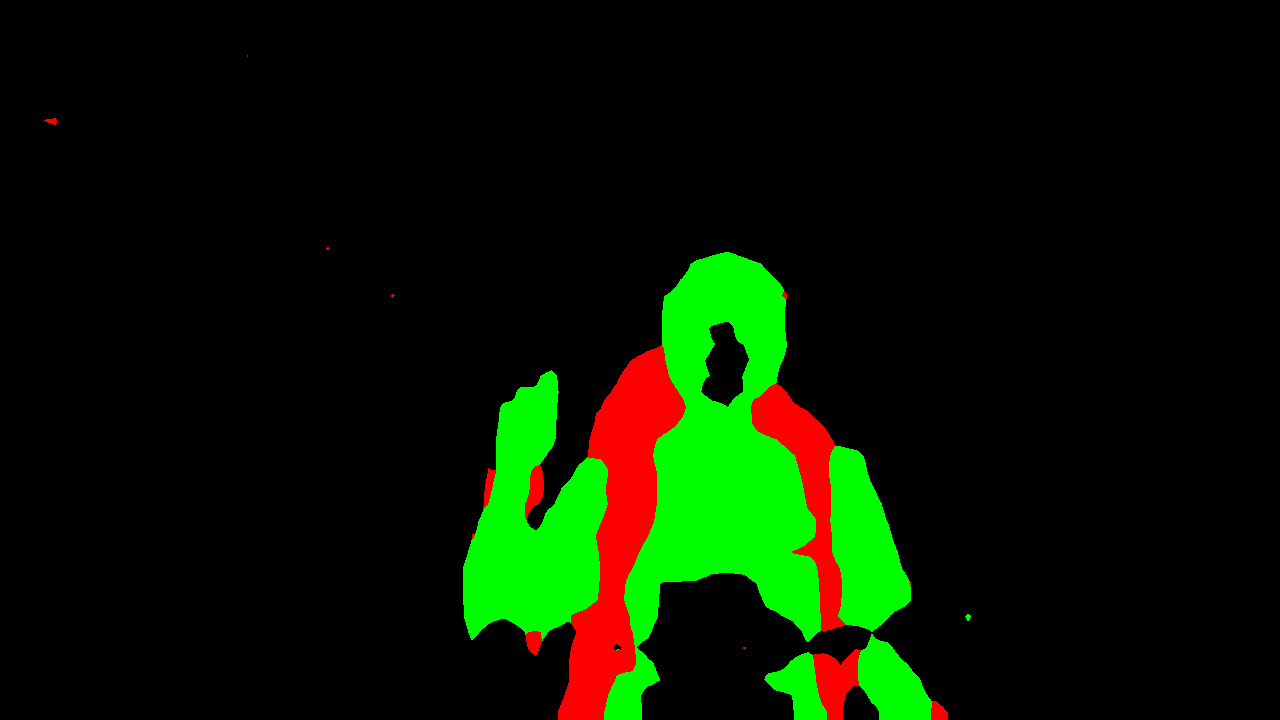}
	
	\includegraphics[width=.32\columnwidth]{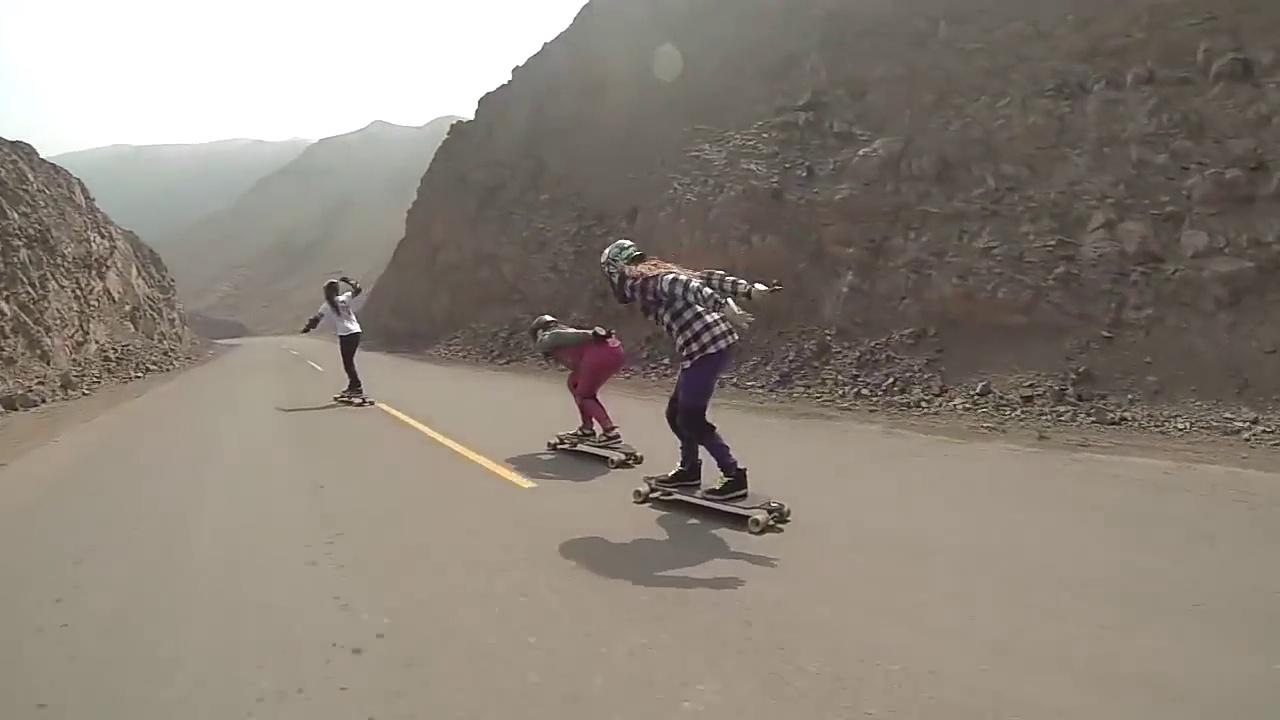}
	\includegraphics[width=.32\columnwidth]{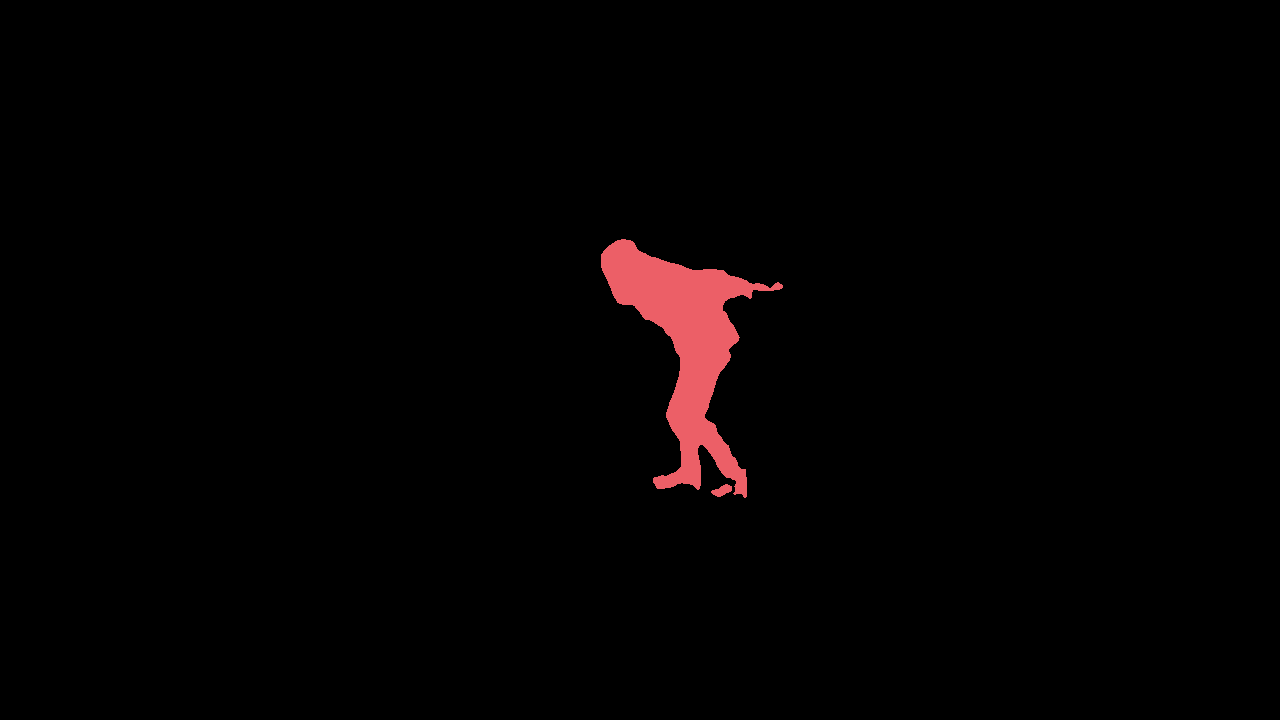}
	\includegraphics[width=.32\columnwidth]{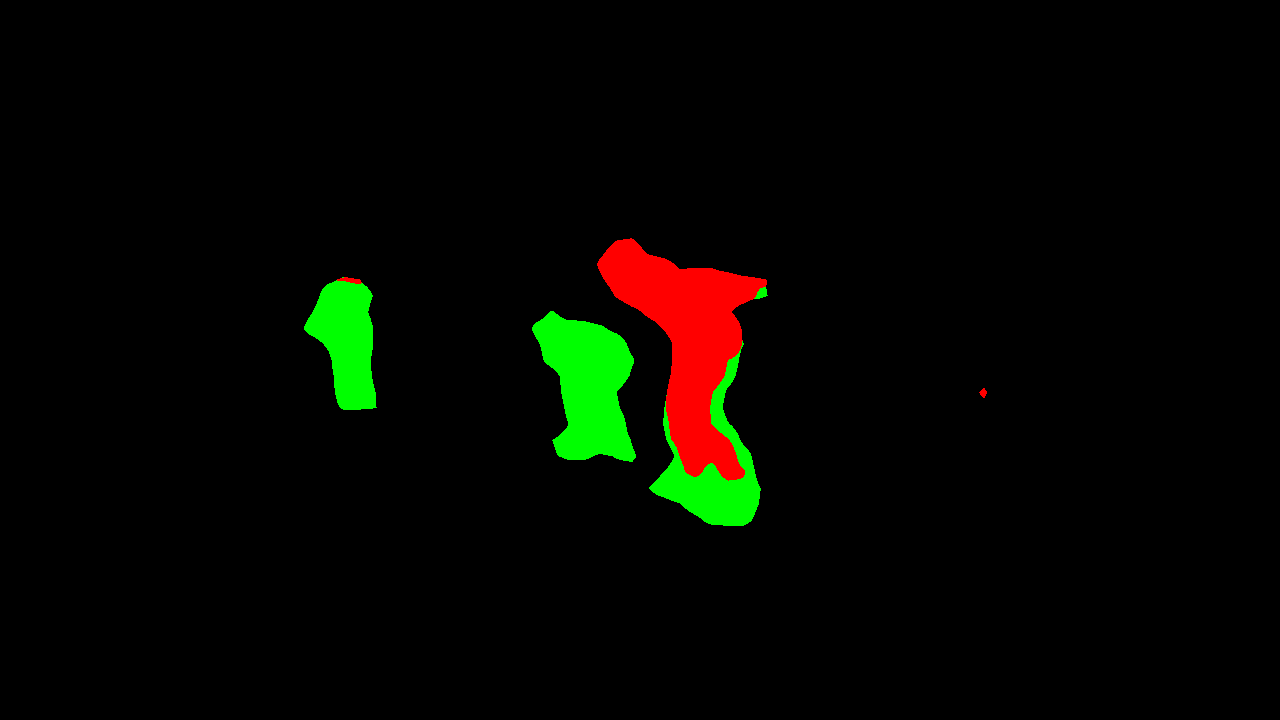}
	
	\includegraphics[width=.32\columnwidth]{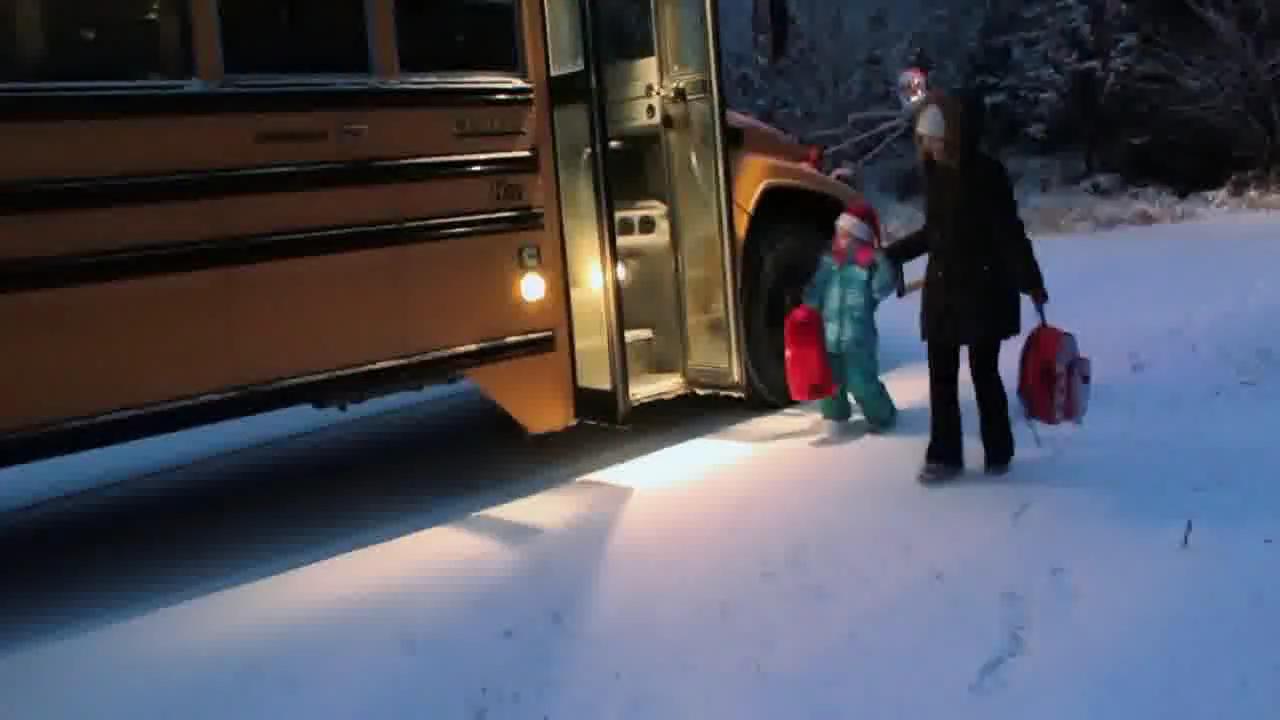}
	\includegraphics[width=.32\columnwidth]{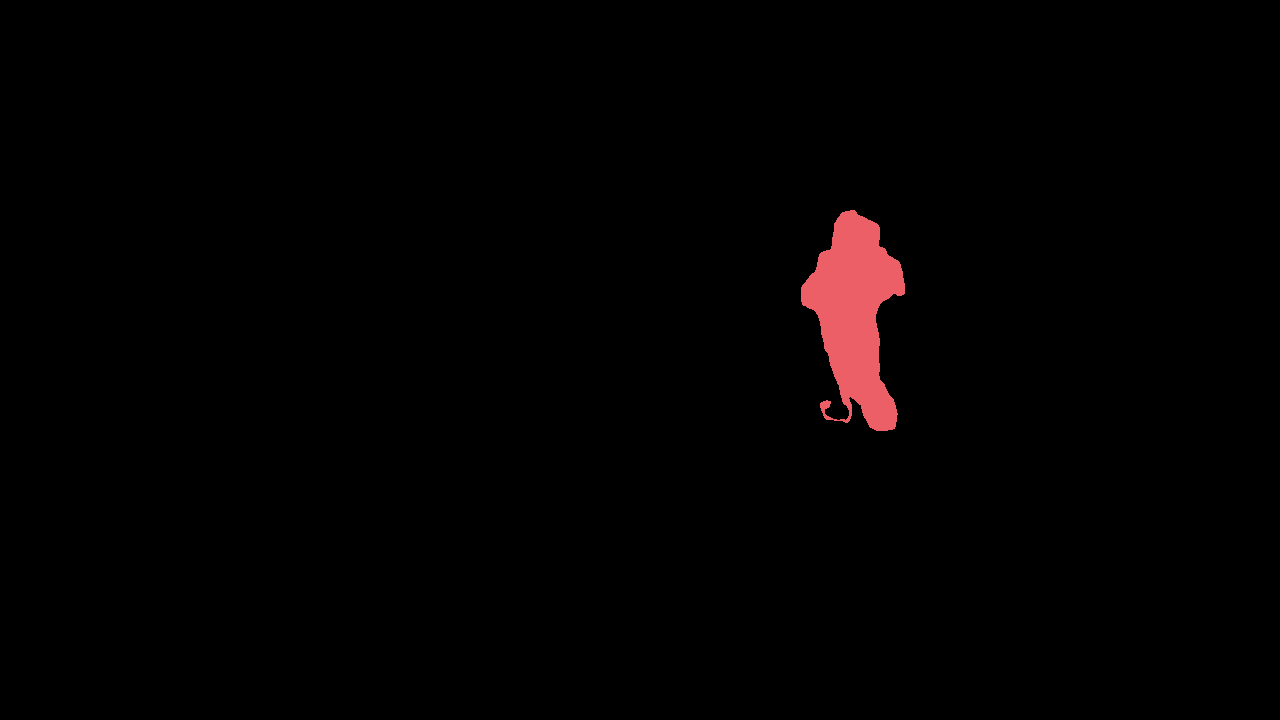}
	\includegraphics[width=.32\columnwidth]{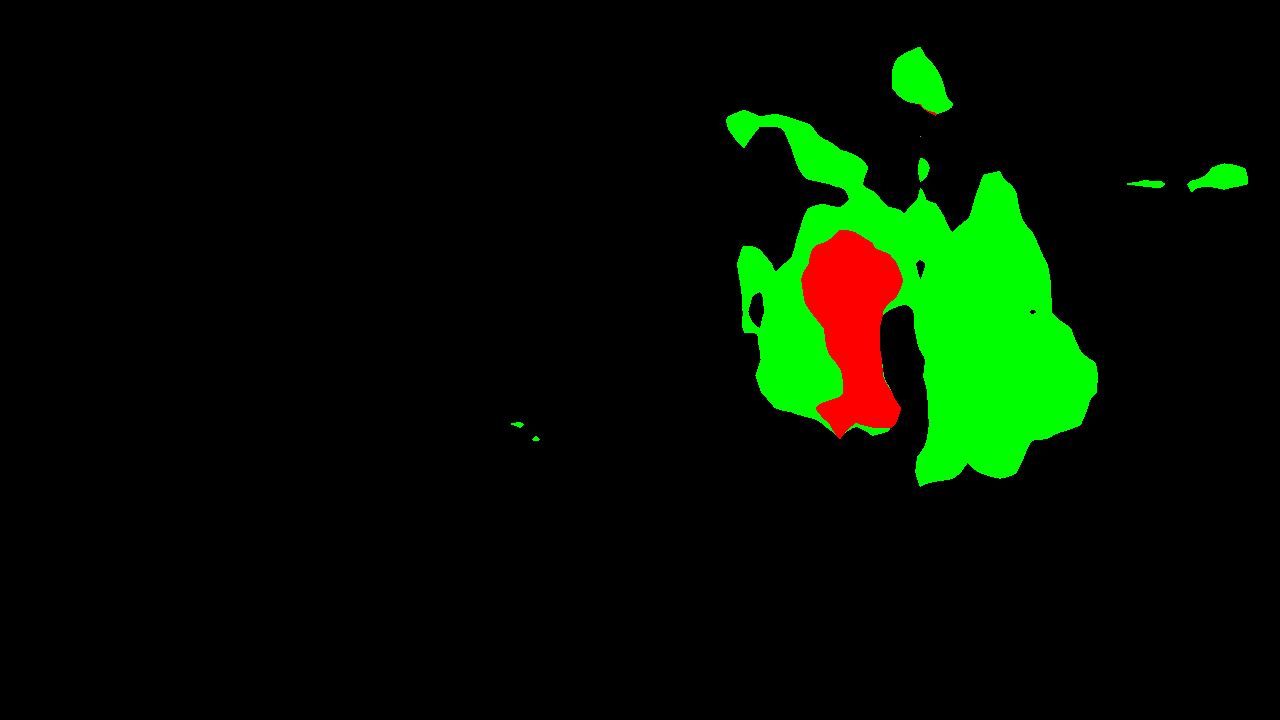}
	
	\includegraphics[width=.32\columnwidth]{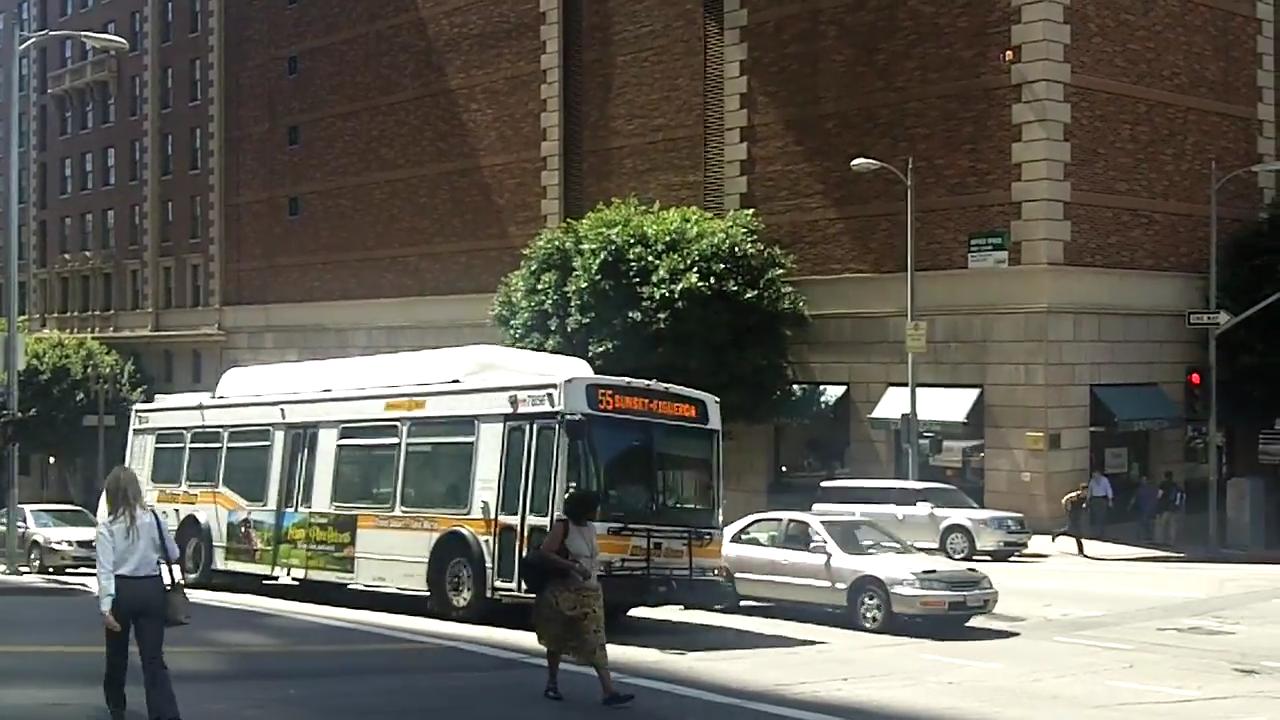}
	\includegraphics[width=.32\columnwidth]{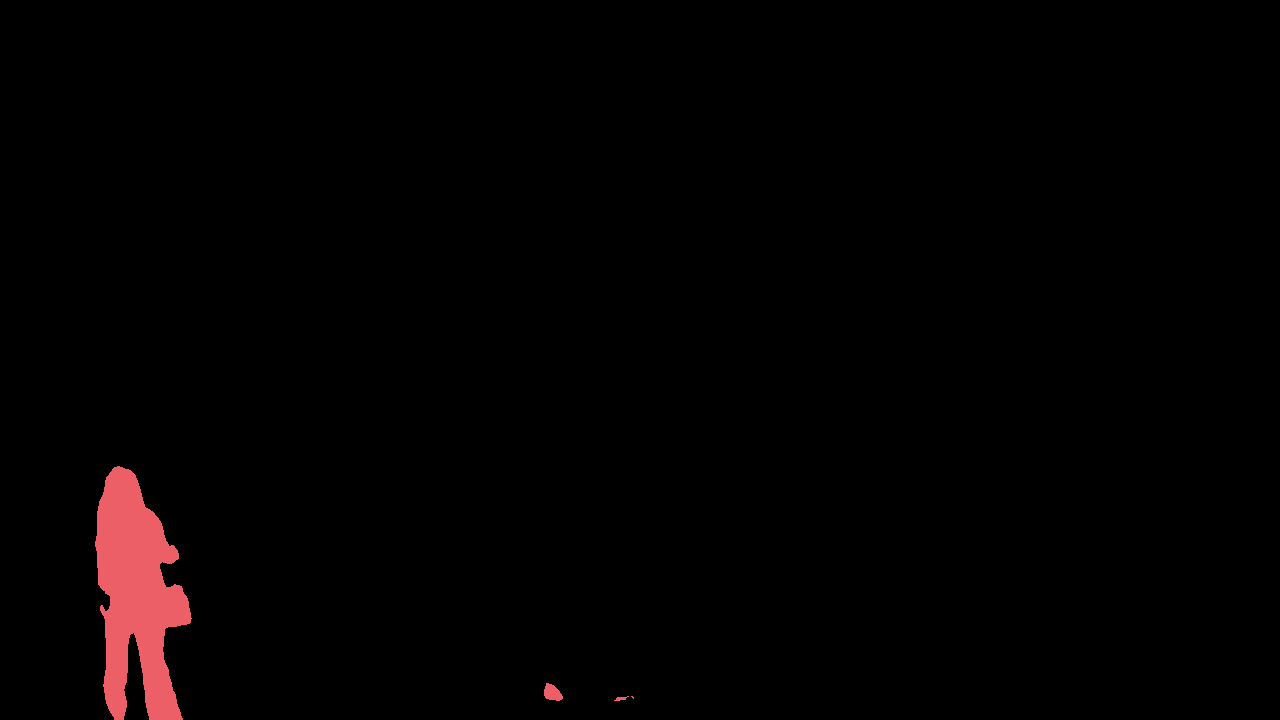}
	\includegraphics[width=.32\columnwidth]{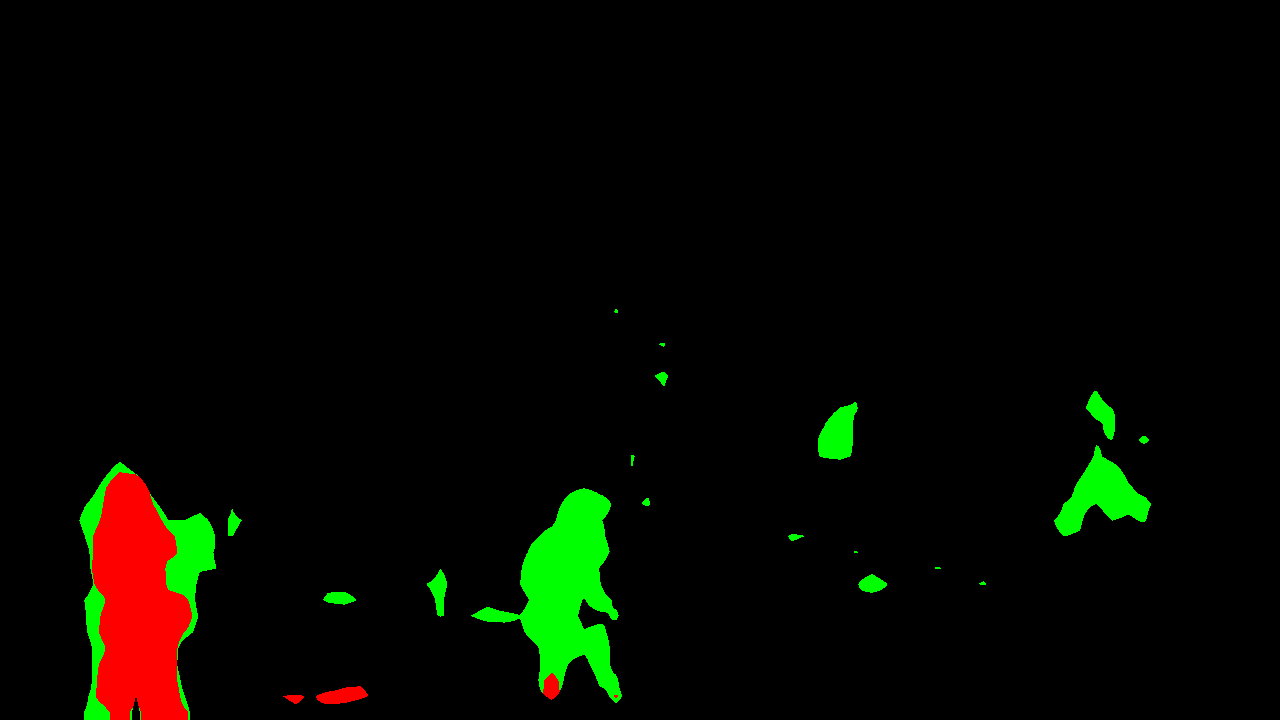}
	\caption{Visualization of the appearance module on five videos from YouTube-VOS. The final segmentation of our approach is shown (middle) together with output of the appearance module (right). The appearance module accurately locates the target (red) with the foreground representation while accentuating potential distractors (green) with the secondary mixture component.}\vspace{-3mm}
	\label{fig:qualitative-lda}
\end{figure}
 
\parsection{Appearance module}
We first analyze the impact of the proposed appearance module (see section~\ref{sec:appearance-module}) by removing it from the network (\emph{No appearance module} in table~\ref{tab:ablation}). This leads to a major reduction in overall performance, from $66.0\%$ to $50.0\%$. The results clearly demonstrate that the introduced appearance module is an essential component in our video object segmentation approach. Further insights are obtained by studying the performance on \emph{seen} and \emph{unseen} classes in table~\ref{tab:ablation}. Note that removing the appearance module causes a $9.1\%$ decrease for classes that are seen during training, and a remarkable $20.6\%$ decrease for unseen classes. Thus, our generative appearance model component is crucial for the generalization to arbitrary objects that are unseen during training. This is explained by the target specific and class-agnostic nature of our appearance module. 

\parsection{Mask-propagation module}
Secondly, we investigate the importance of the mask-propagation module (see section~\ref{sec:architecture}). 
Refraining from propagating the mask predicted in the previous frame (\emph{No mask-prop module} in table~\ref{tab:ablation}) results in a $2.0\%$ reduction in performance. While this reduction is significant, the importance of the mask-propagation module is small compared to that of the appearance module.

\parsection{Gaussian mixture components}
As described in section~\ref{sec:appearance-module}, we employ two Gaussian mixture components to model the foreground and background, respectively. In addition to the base mixture component, a secondary Gaussian mixture component is added to capture hard examples that are not accurately modelled by a unimodal distribution. We investigate the impact of this additional mixture components by removing them from our model. The resulting version (\emph{Unimodal appearance} in table~\ref{tab:ablation}) thus only employs a single base mixture component for each class. The resulting performance drop of $1.6\%$ indicates the importance of modeling hard examples in the presence of distractor objects. 

The impact of the multi-modal generative model is also analyzed qualitatively in fig.~\ref{fig:qualitative-lda}. The mixture component dedicated to hard negative image regions is able to model other objects in the vicinity of the target (row 1 and 2) and accurately captures other objects of the same class (row 3-5). Note that both the appearance module's output and the final segmentations are soft, and only for the purpose of visualization we show the arguments of the maxima.

\parsection{Model update}
We investigate the impact of updating the generative model in each frame using \eqref{eq:param-update}. The version \emph{No update} (table~\ref{tab:ablation}) only uses the initial frame to compute the mixture model parameters \eqref{eq:gen-param}, and no update \eqref{eq:param-update} is performed during training and inference. Updating the generative model to capture changes in the target and background appearance leads to a $1.1\%$ improvement in performance.

\parsection{Appearance module output}
As previously described, our appearance module outputs the log-probability scores \eqref{eq:scores}. To validate this choice, we also compare with outputting the posterior probabilities (\emph{Appearance SoftMax} in table~\ref{tab:ablation}), obtained by adding a \verb|SoftMax| layer after computing the scores \eqref{eq:scores}, between the appearance and fusion modules. This leads to a significant degradation in performance ($-10.2\%$).

These results are in line with conventional techniques in segmentation \cite{FCN} and classification \cite{Alexnet}, where activations in the network are not converted to probabilities until the final output layer.

\begin{table}[!t]
	\centering
	\resizebox{\columnwidth}{!}{%
		\begin{tabular}{l|c|c|c|c}
			Method & O-Ft & $\mathcal{G}$ overall (\%) & $\mathcal{J}$ seen (\%) & $\mathcal{J}$ unseen (\%)\\\hline
			S2S \cite{YTVOS-extra} & \yesmark & \first{64.4} & \first{71.0} & \first{55.5}\\
			OSVOS \cite{OSVOS} & \yesmark & 58.8 & 59.8 & 54.2\\
			OnAVOS \cite{OnAVOS} & \yesmark & 55.2 & 60.1 & 46.6\\
			MSK \cite{MSK} & \yesmark & 53.1 & 59.9 & 45.0\\
			\hline
			OSMN \cite{OSMN} & \nomark & 51.2 & 60.0 & 40.6\\
			S2S \cite{YTVOS-extra} & \nomark & 57.6 & 66.7 & 48.2\\
			RGMP \cite{RGMP} & \nomark & 53.8 & 59.5 & 45.2\\
			\textbf{Ours} & \nomark & \first{66.0} & \first{66.9} & \first{61.2}\\
		\end{tabular}
	}\vspace{1mm}
	\caption{State-of-the-art comparison on the YouTubeVOS benchmark. Our approach obtains the best overall performance ($\mathcal{G}$) despite not performing any online fine-tuning (O-Ft). Further, our approach provides a large gain in performance for categories unseen during training ($\mathcal{J}$ unseen), compared to existing methods.}\vspace{-3mm}
	\label{tab:sota-ytvos}
\end{table}

\parsection{End-to-end learning}
Finally, we analyze the impact of end-to-end differentiation and training in our approach. Specifically, we investigate the importance of end-to-end differentiability in the learning stage of the appearance module. The comparison is performed by not backpropagating through the model inference computation \eqref{eq:gen-param} during the training of the network. Note that, the rest of the framework remains unchanged. The resulting method (\emph{No end-to-end}  in table~\ref{tab:ablation}) obtains poor results, with a total degradation of $7.2\%$ in overall performance. This highlights the importance of permitting \emph{true} end-to-end learning.

\subsection{State-of-the-Art Comparison}
We compare our approach with the state-of-the-art on three video object segmentation benchmarks: YouTube-VOS \cite{YTVOS}, DAVIS2017 \cite{DAVIS17}, and DAVIS2016 \cite{DAVIS16}.

\begin{table}[!t]
  \centering
  \resizebox{.6\columnwidth}{!}{%
  \begin{tabular}{l|c|c|c}
    Method & O-Ft & Causal & $\mathcal{J}$ (\%)\\\hline
    CINM \cite{CINM} & \yesmark & \yesmark & \first{67.2}\\
    OSVOS-S \cite{OSVOS-S} & \yesmark & \yesmark & 64.7\\
    OnAVOS \cite{OnAVOS} & \yesmark & \yesmark & 61.6\\
    OSVOS \cite{OSVOS} & \yesmark & \yesmark & 56.6\\
    \hline
    DyeNet \cite{DyeNet} & \nomark & \nomark & \first{67.3}\\
    \hline
    RGMP \cite{RGMP} & \nomark & \yesmark & 64.8\\
    VM \cite{VM} & \nomark & \yesmark & 56.5\\
    FAVOS \cite{FAVOS} & \nomark & \yesmark & 54.6\\
    OSMN \cite{OSMN} & \nomark & \yesmark & 52.5\\
    \textbf{Ours} & \nomark & \yesmark & \first{67.2}\\
  \end{tabular}
  }\vspace{1mm}
  \caption{State-of-the-art comparison on the DAVIS2017 validation set. For each method we report whether it employs online fine-tuning (O-Ft), is causal, and the final performance $\mathcal{J}$ (\%). Our approach obtains superior results compared to state-of-the-art methods without online fine-tuning. Further, our approach closes the performance gap to existing methods employing online fine-tuning.}\vspace{-3mm}
  \label{tab:sota-davis17}
\end{table}

\begin{table}
  \centering
  \resizebox{.7\columnwidth}{!}{%
  \begin{tabular}{l|c|c|r|c}
  Method & O-Ft & Causal & Speed & $\mathcal{J}$ (\%)\\\hline
  OnAVOS \cite{OnAVOS} & \yesmark & \yesmark & 13s & 86.1\\
  OSVOS-S \cite{OSVOS-S} & \yesmark & \yesmark & 4.5s & 85.6\\
  MGCRN \cite{MGCRN} & \yesmark & \yesmark & 0.73s & 84.4\\
  CINM \cite{CINM} & \yesmark & \yesmark & \textgreater30s & 83.4\\
  LSE \cite{LSE} & \yesmark & \yesmark & & 82.9\\

  OSVOS \cite{OSVOS} & \yesmark & \yesmark & 9s & 79.8\\
  MSK \cite{MSK} & \yesmark & \yesmark & 12s & 79.7\\
  SFL \cite{SFL} & \yesmark & \yesmark & 7.9s & 74.8\\

  \hline
  DyeNet \cite{DyeNet} & \nomark & \nomark & 0.42s & 84.7\\
  \hline
  FAVOS \cite{FAVOS} & \nomark & \yesmark & 1.80s & 82.4\\
  RGMP \cite{RGMP} & \nomark & \yesmark & 0.13s & 81.5\\
  VM \cite{VM} & \nomark & \yesmark & 0.32s & 81.0\\
  MGCRN \cite{MGCRN} & \nomark & \yesmark & 0.36s & 76.4\\

  PML \cite{PML} & \nomark & \yesmark & 0.28s & 75.5\\
  OSMN \cite{OSMN} & \nomark & \yesmark & 0.14s & 74.0\\
  CTN \cite{CTN} & \nomark & \yesmark & 1.30s & 73.5\\
  VPN \cite{VPN} & \nomark & \yesmark & 0.63s & 70.2\\
  MSK \cite{MSK} & \nomark & \yesmark & 0.15s & 69.9\\
  \textbf{Ours} & \nomark & \yesmark & 0.07s & 82.0\\
  \end{tabular}
  }
  \caption{State-of-the-art comparison on DAVIS2016 validation set, which is a subset of DAVIS2017. For each method we report whether it employs online fine-tuning (O-Ft), is causal, the computation time (if available), and the final performance $\mathcal{J}$ (\%). Our approach obtains competitive results compared to causal methods without online fine-tuning.}\vspace{-3mm}
  \label{tab:sota-davis16}
\end{table}

\begin{figure*}[!t]
  \centering
  \parbox{.14\textwidth}{\centering Image}
  \parbox{.14\textwidth}{\centering Ground Truth}
  \parbox{.14\textwidth}{\centering RGMP \cite{RGMP}}
  \parbox{.14\textwidth}{\centering CINM \cite{CINM}}
  \parbox{.14\textwidth}{\centering FAVOS \cite{FAVOS}}
  \parbox{.14\textwidth}{\centering Ours}

  \includegraphics[width=.14\textwidth]{images/im_bmx-trees_00015}
  \includegraphics[width=.14\textwidth]{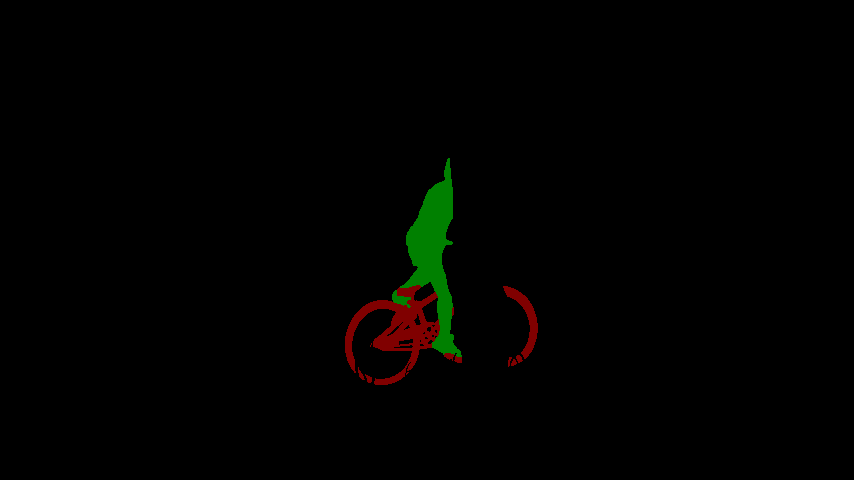}
  \includegraphics[width=.14\textwidth]{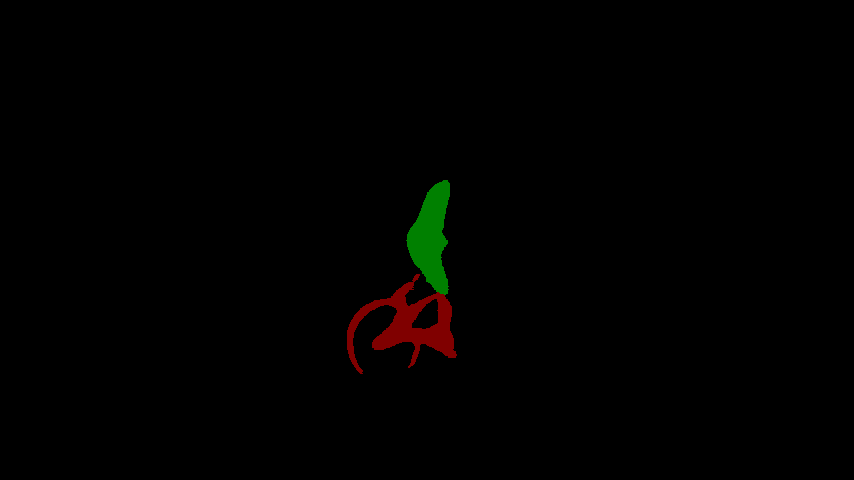}
  \includegraphics[width=.14\textwidth]{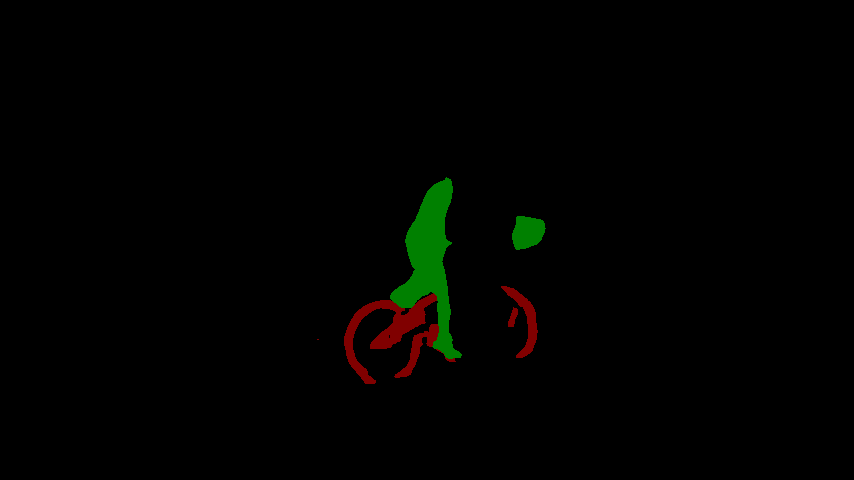}
  \includegraphics[width=.14\textwidth]{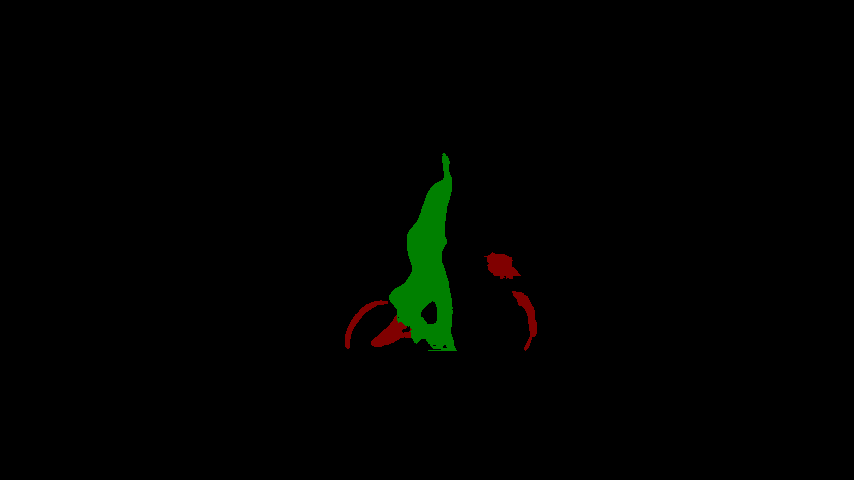}
  \includegraphics[width=.14\textwidth]{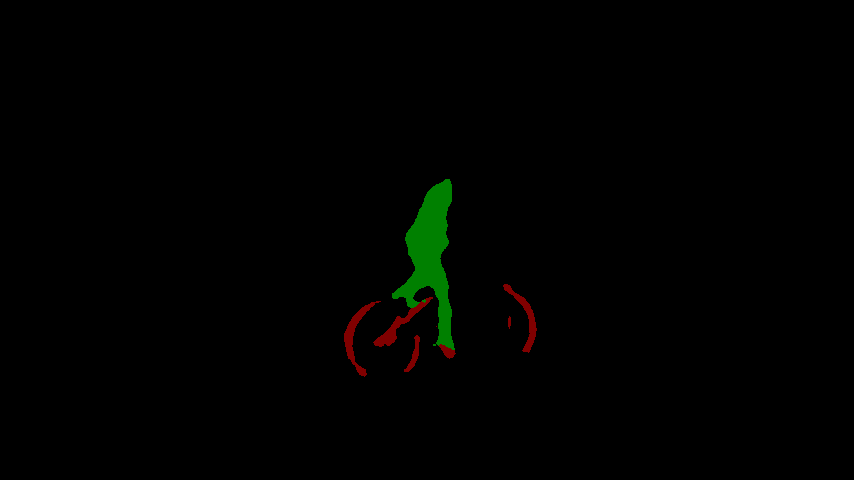}
  
  \includegraphics[width=.14\textwidth]{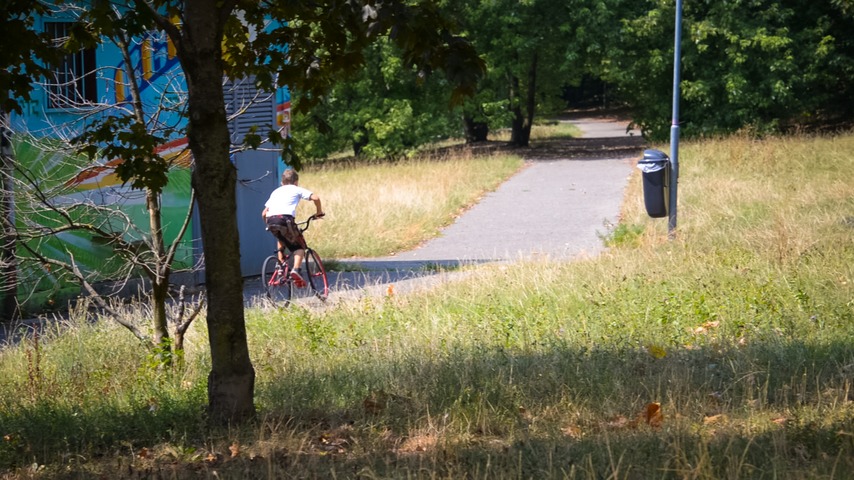}
  \includegraphics[width=.14\textwidth]{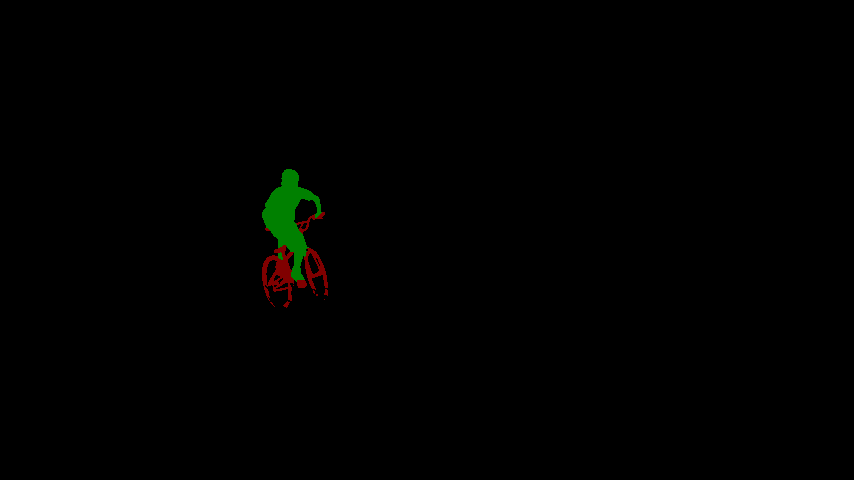}
  \includegraphics[width=.14\textwidth]{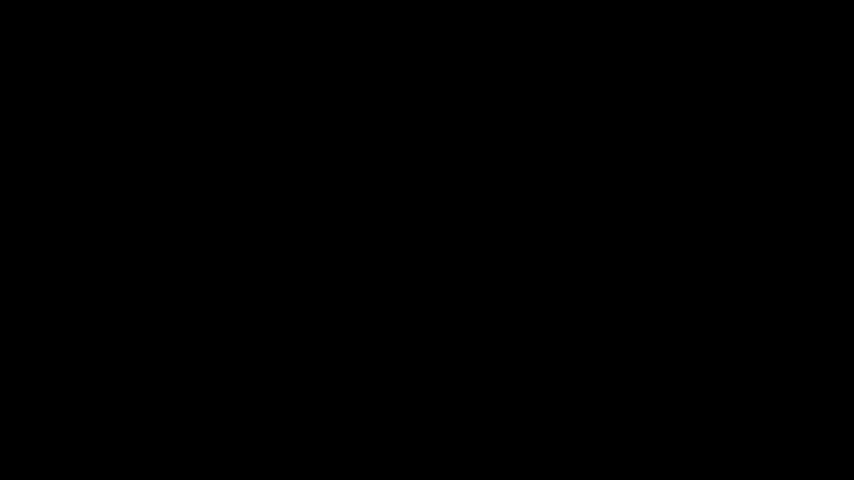}
  \includegraphics[width=.14\textwidth]{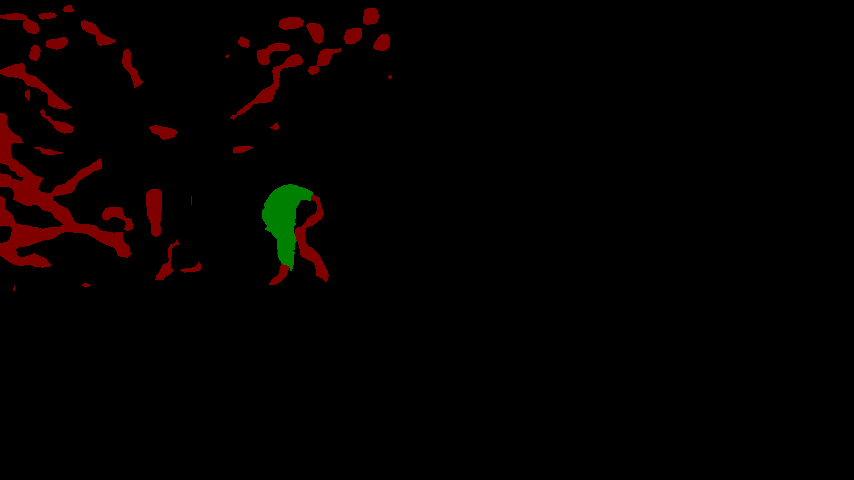}
  \includegraphics[width=.14\textwidth]{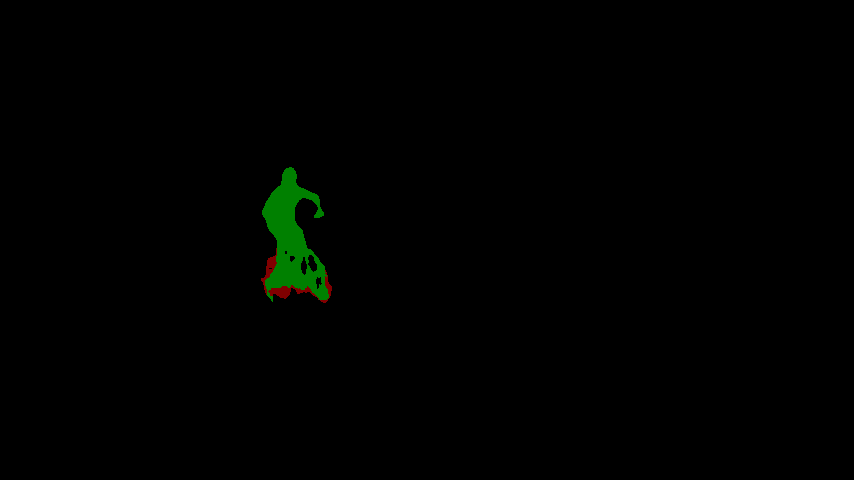}
  \includegraphics[width=.14\textwidth]{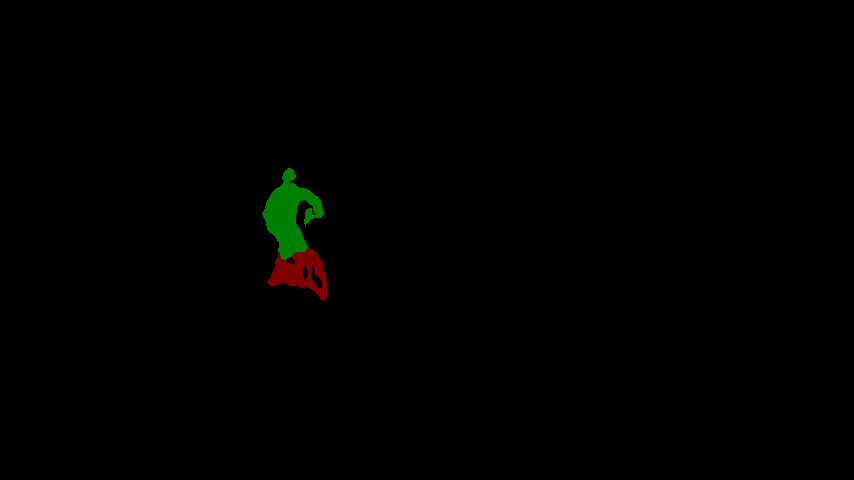}
  
  \includegraphics[width=.14\textwidth]{images/im_judo_00011}
  \includegraphics[width=.14\textwidth]{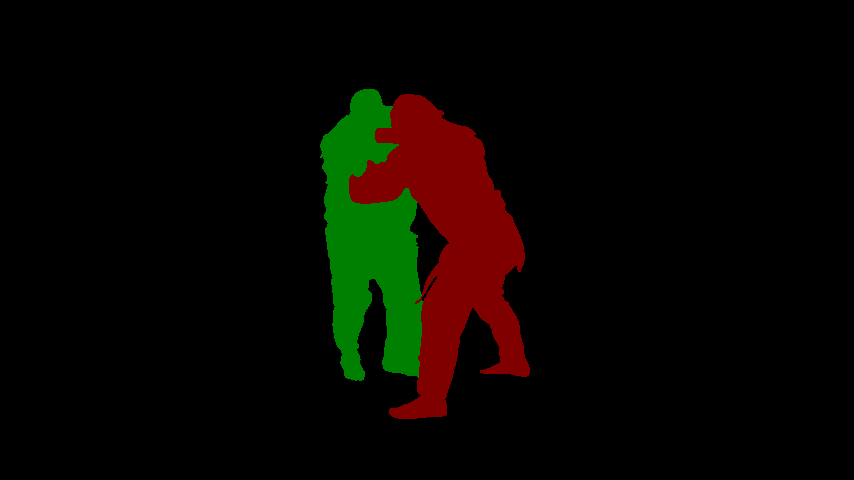}
  \includegraphics[width=.14\textwidth]{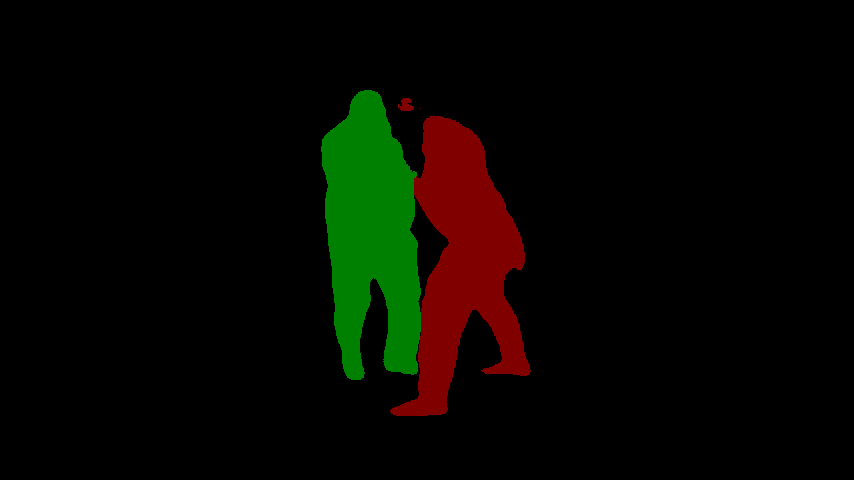}
  \includegraphics[width=.14\textwidth]{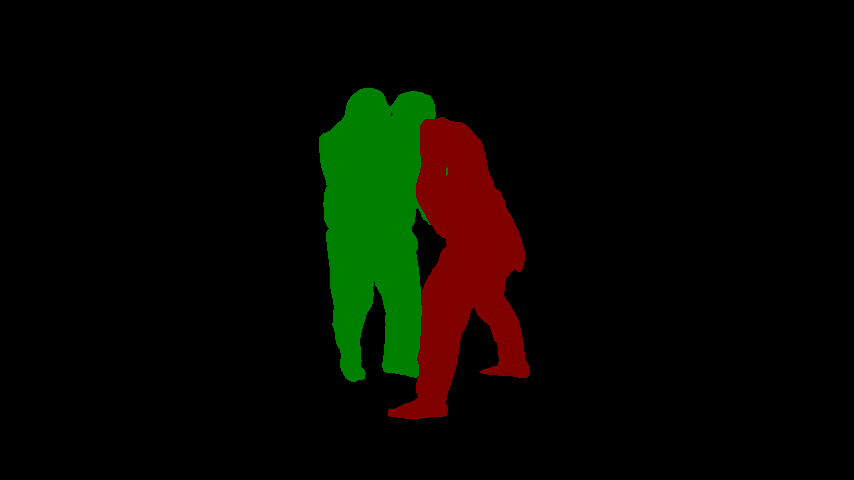}
  \includegraphics[width=.14\textwidth]{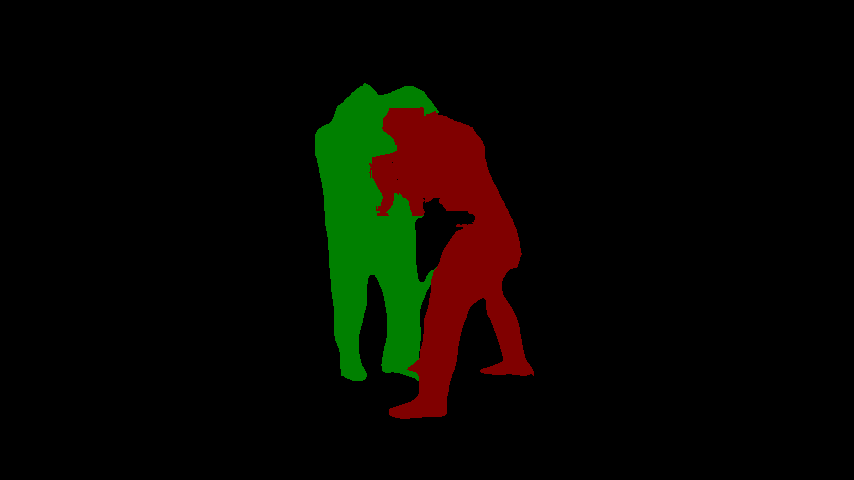}
  \includegraphics[width=.14\textwidth]{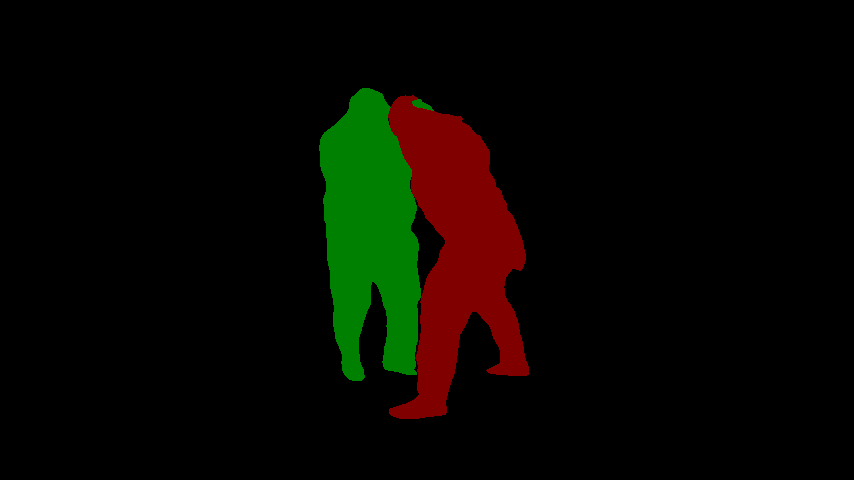}
  
  \includegraphics[width=.14\textwidth]{images/im_judo_00021}
  \includegraphics[width=.14\textwidth]{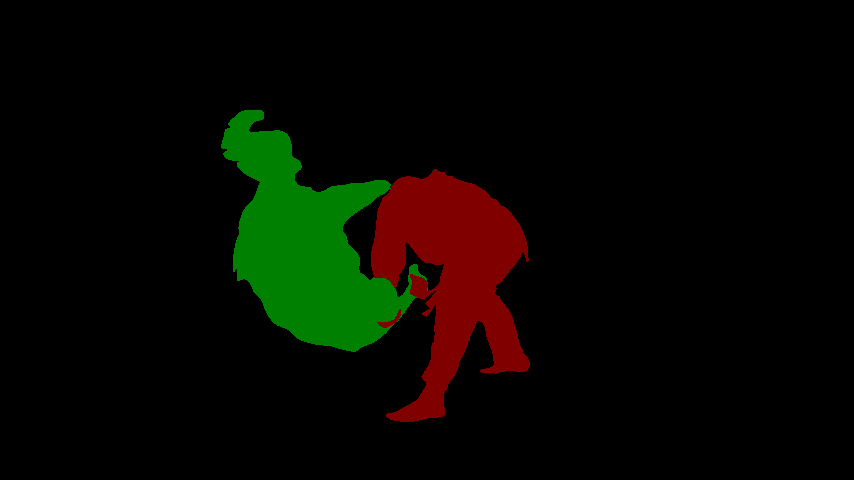}
  \includegraphics[width=.14\textwidth]{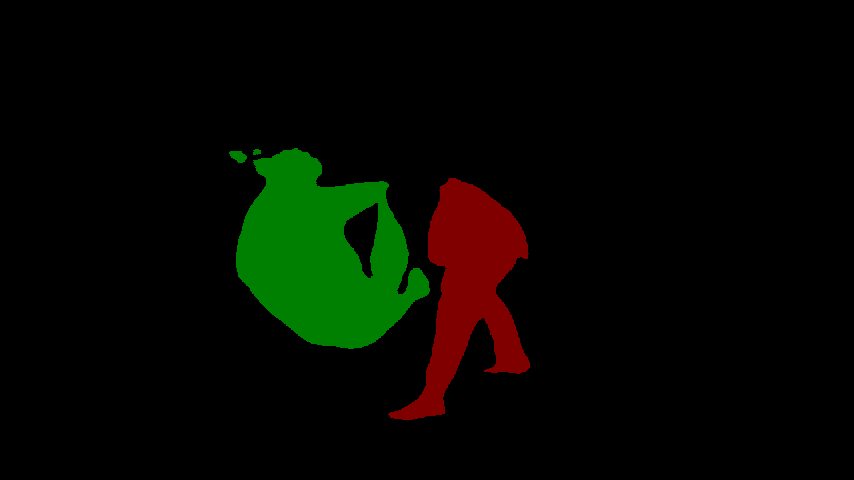}
  \includegraphics[width=.14\textwidth]{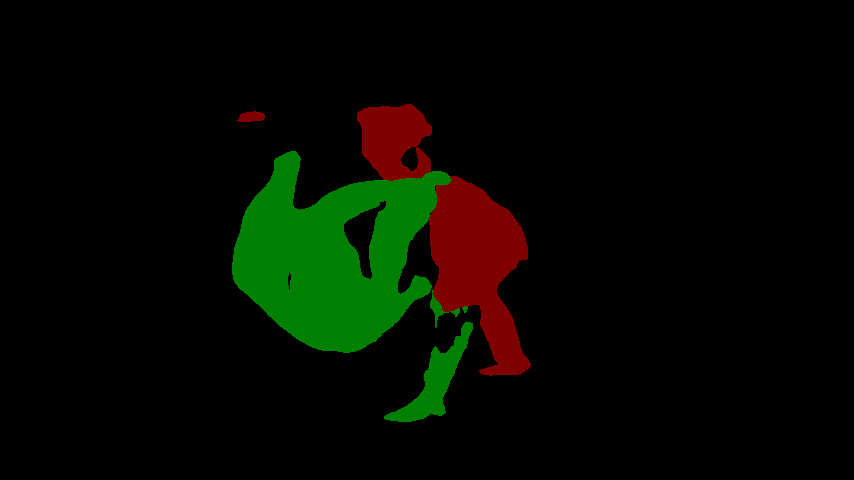}
  \includegraphics[width=.14\textwidth]{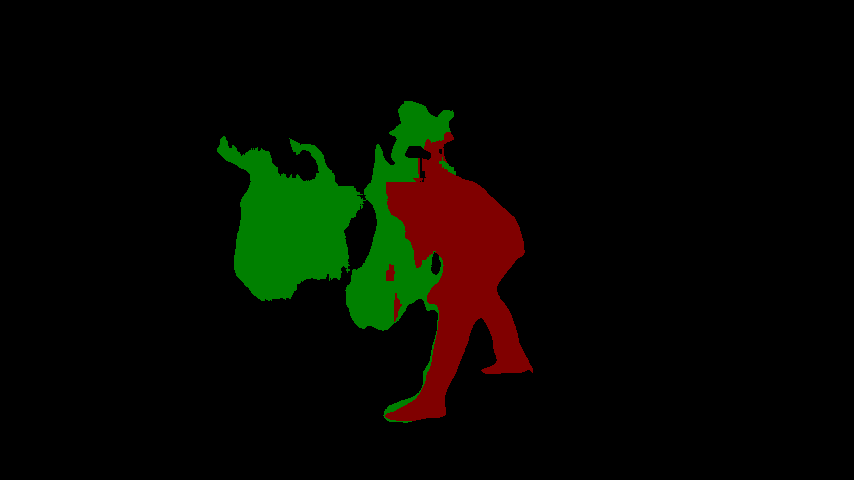}
  \includegraphics[width=.14\textwidth]{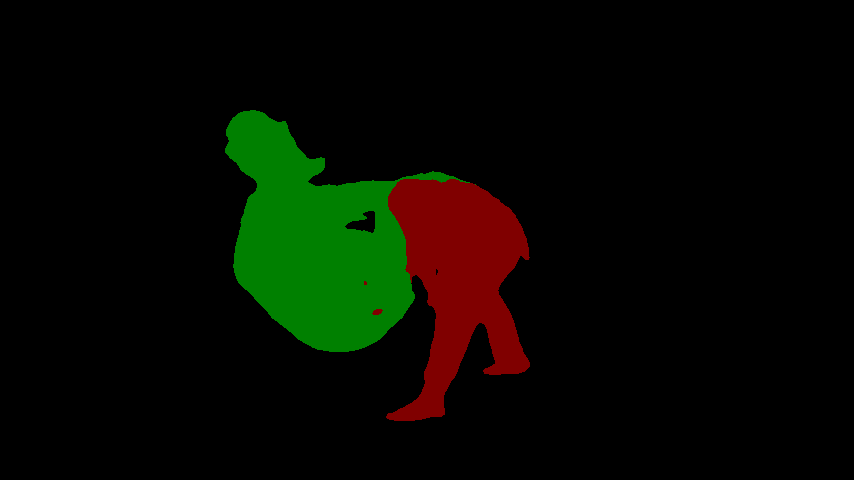}
  
  \includegraphics[width=.14\textwidth]{images/im_bike-packing_00008}
  \includegraphics[width=.14\textwidth]{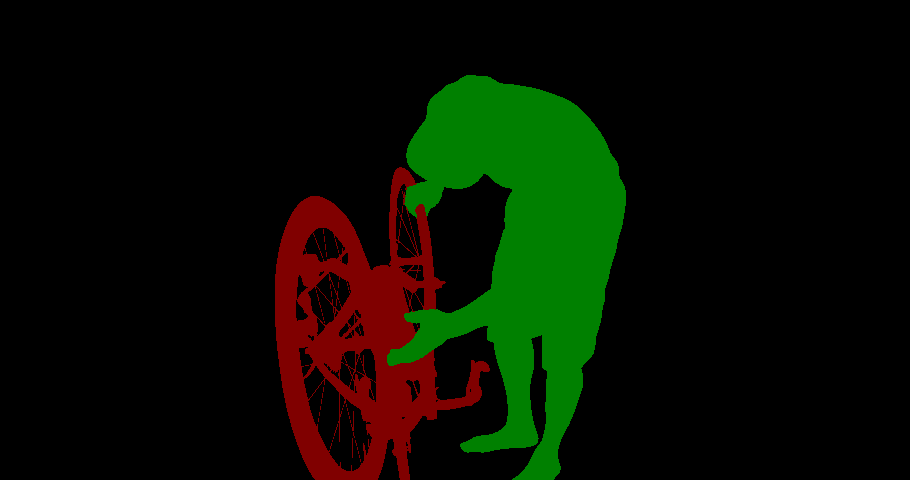}
  \includegraphics[width=.14\textwidth]{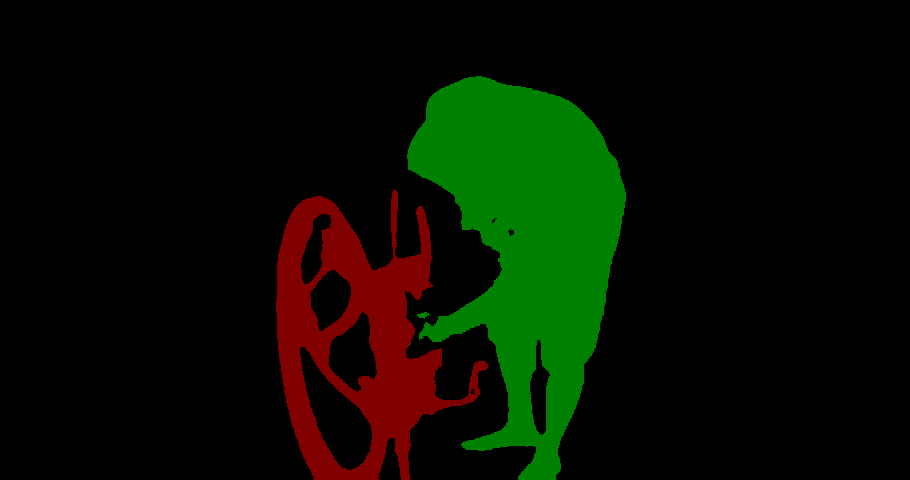}
  \includegraphics[width=.14\textwidth]{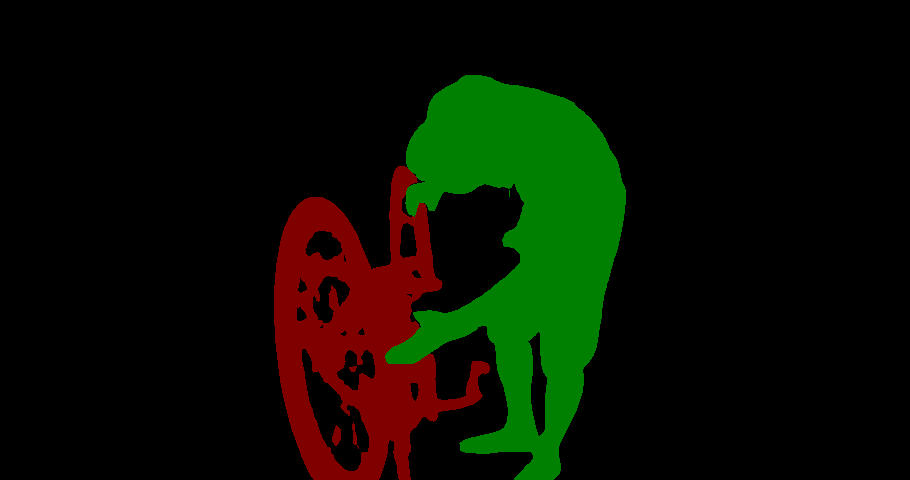}
  \includegraphics[width=.14\textwidth]{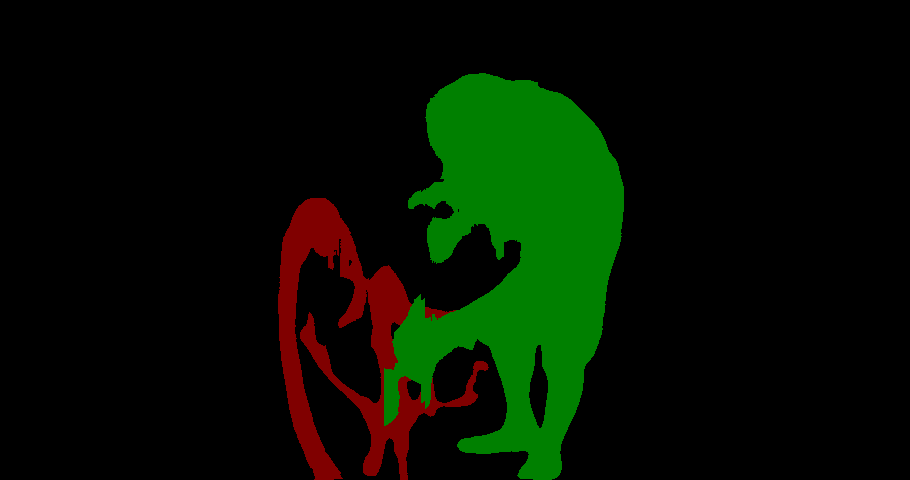}
  \includegraphics[width=.14\textwidth]{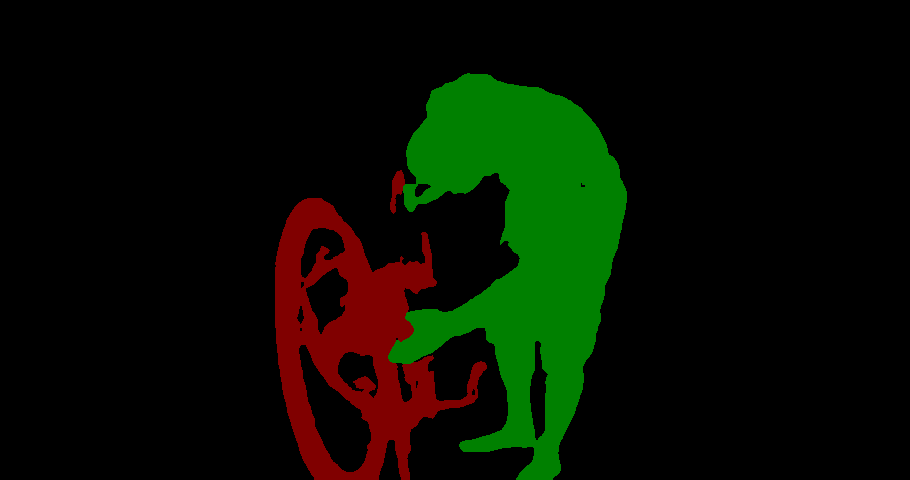}
  
  \includegraphics[width=.14\textwidth]{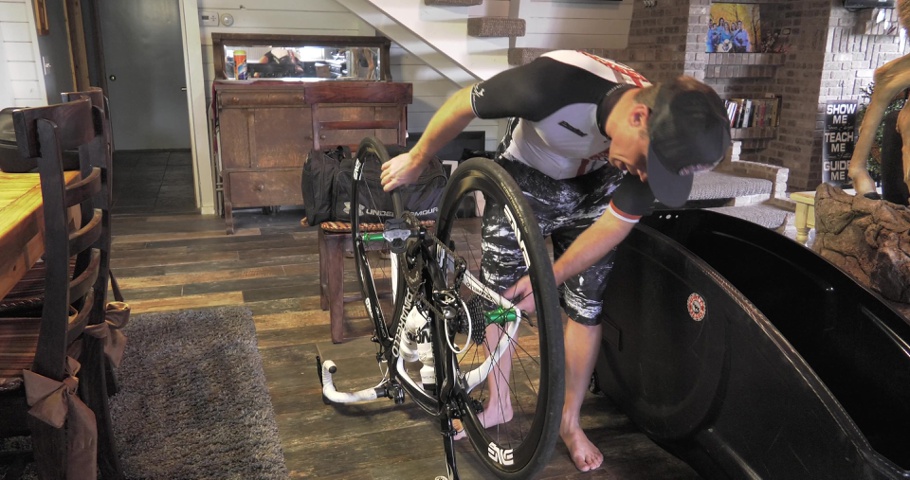}
  \includegraphics[width=.14\textwidth]{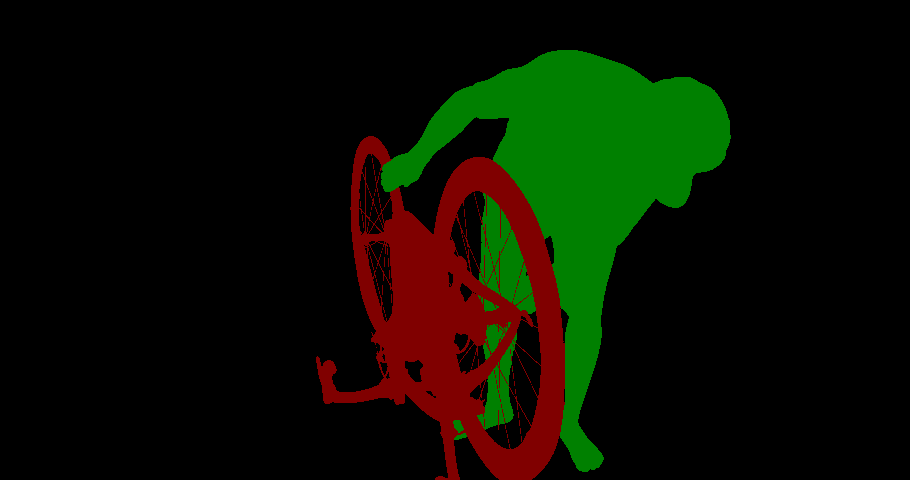}
  \includegraphics[width=.14\textwidth]{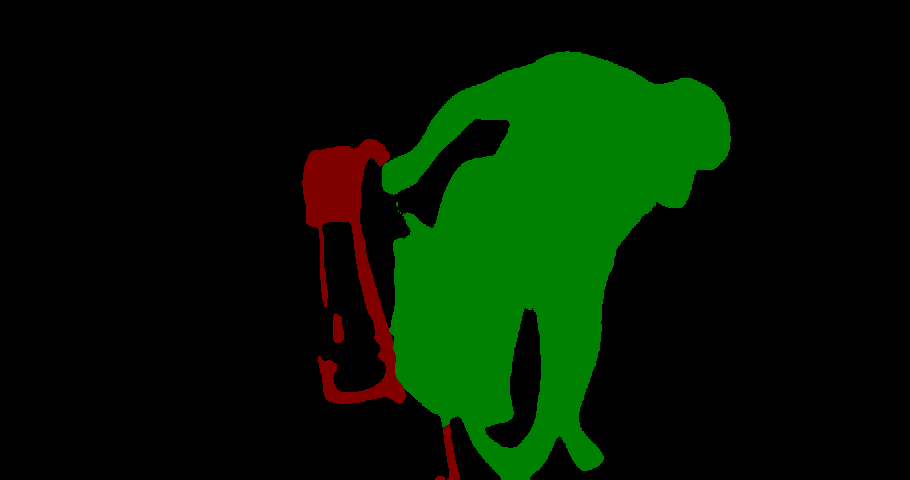}
  \includegraphics[width=.14\textwidth]{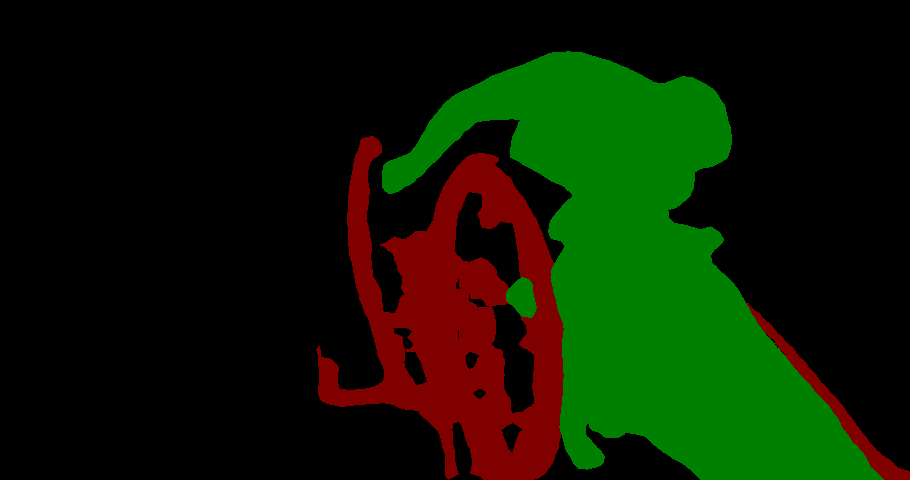}
  \includegraphics[width=.14\textwidth]{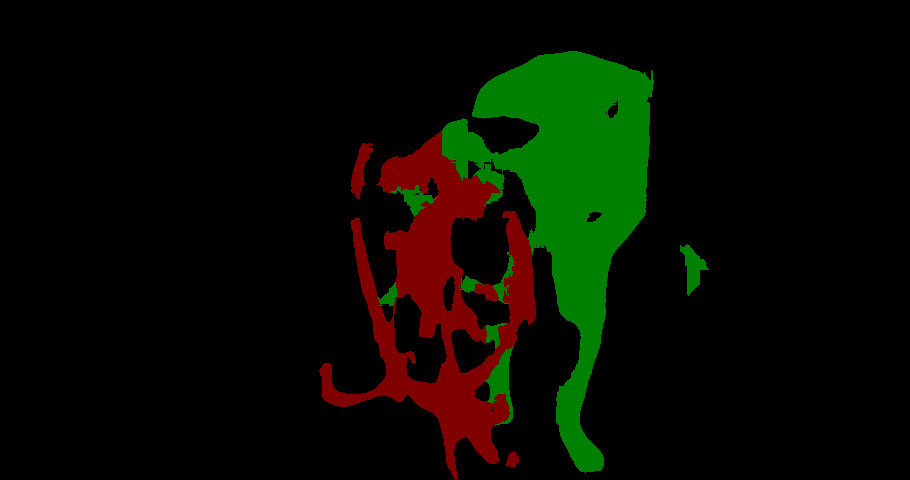}
  \includegraphics[width=.14\textwidth]{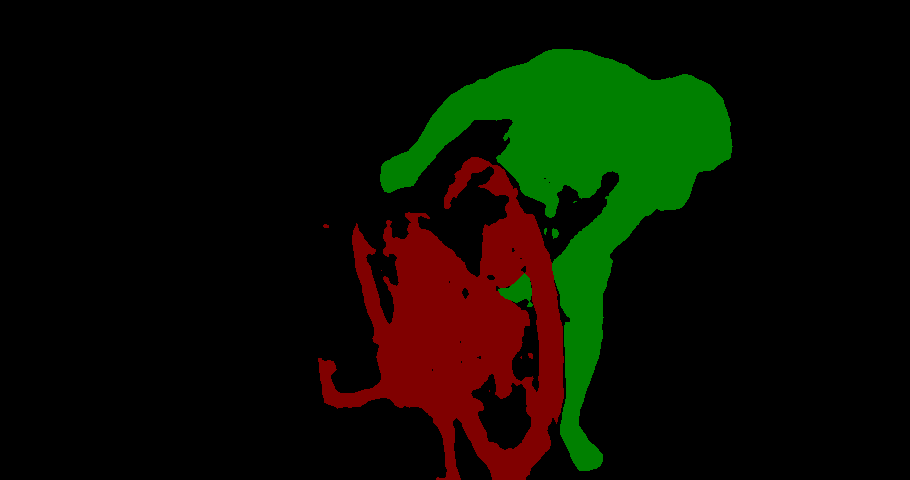}

  \caption{Qualitative comparison between our approach and 3 state-of-the-art approaches. Our approach is able to accurately segment all targets, demonstrating robustness to occlusions and successfully discriminating between different objects. This is largely thanks to the powerful appearance model in our architecture.}
  \vspace{-.22cm}

  \label{fig:qualitative-comparison}
\end{figure*}

\parsection{YouTube-VOS}
This recently introduced large-scale dataset contains 474 sequences with 91 classes, 26 of which are not included in the YouTube-VOS training set. We use the official validation set, as in section~\ref{sec:ablation}. We compare our approach with all, to our best knowledge, published results \cite{YTVOS}. Additionally, we evaluate the RGMP method, using the code provided by the authors. The results are shown in table~\ref{tab:sota-ytvos}. For each approach, we indicate if the method employs online fine-tuning (O-Ft) and if it is causal, i.e.\ if the segmentation output depends on future frames in the video. Here we let \yesmark and \nomark indicate yes and no, respectively. Among previous approaches performing extensive online fine-tuning in the first frame, OSVOS and OnAVOS achieve final scores of $58.8\%$ and $55.2\%$. For the S2S method, we compare with two versions: one with and one without online fine-tuning, obtaining $64.4\%$ and $57.6\%$, respectively.
Our approach obtains a final score of $66.0\%$, significantly outperforming state-of-the-art without invoking any online fine-tuning. Furthermore, our method performs notably well on the \emph{unseen} category, which only considers objects that are not seen during training. Again, this demonstrates the effectiveness of our class-agnostic appearance module, which generalizes to arbitrary target objects.

\parsection{DAVIS2017}
The dataset comprises 30 videos with one or multiple target objects. The results are shown in table~\ref{tab:sota-davis17}. Among existing methods, DyeNet is the only approach that is non-causal, since it processes the entire video in a bidirectional manner. It is therefore not applicable to real-time or online systems. 
The RGMP method, achieving a score of $64.8\%$, relies on mask propagation and an appearance model constructed by simply concatenating image features from the first frame. VideoMatch (VM) stores foreground and background feature vectors that are then matched with feature vectors in the test image. This method obtains a final result of $56.5\%$. The proposed method, employing an end-to-end differentiable generative probabilistic appearance model, achieves a score of $67.2\%$. Our approach outperforms all causal methods not invoking online fine-tuning, and is even on par with the best non-causal and online fine-tuning-based techniques. 

\parsection{DAVIS2016} 
For completeness, we also evaluate our approach on DAVIS2016. It is a subset of DAVIS2017, containing 20 videos labeled with a single object. The small size and number of objects in DAVIS2016 limits the diversity. It has therefore become highly saturated over the years. In table~\ref{tab:sota-davis16} we show the final result of each method, along with computational time reported by the respective authors. Our approach obtains a competitive performance of $82.0\%$ compared to state-of-the-art. Unlike our method, the top performing approaches on DAVIS2016, such as OSVOS, OnAVOS, and FAVOS do not generalize well to the larger and more diverse YouTube-VOS and DAVIS2017 datasets. 

\subsection{Qualitative Evaluation}
We qualitatively compare our approach with three state-of-the-art approaches (RGMP \cite{RGMP}, CINM \cite{CINM}, FAVOS \cite{FAVOS}) on three videos from DAVIS2017. The results are shown in fig.~\ref{fig:qualitative-comparison}. RGMP tends to lose parts of objects, and struggles with discrimination between different objects. While CINM can produce detailed segmentation masks (row 5), it suffers from several failure modes (row 2, 4, 6). FAVOS struggles with discriminating targets (row 2, 6) and fails to capture details (row 6) or precise boundaries (row 4). The proposed approach succeeds to accurately segment both targets in all scenarios while being one or several orders of magnitude faster compared to FAVOS and CINM, respectively.

%% file: conclusion.tex
\section{Conclusion}
We propose to address the VOS problem by learning the appearance of the target in an efficient and differentiable manner, avoiding the drawbacks of existing matching or online-finetuning based approaches. The target appearance is modelled as a mixture of Gaussians in an embedding space, and we show that both learning and inference of this model can be expressed in closed form. This permits the implementation of the appearance model as a component in a neural network that is trained on end-to-end. We thoroughly analyze the proposed approach and demonstrate its effectiveness on three benchmarks, resulting in state-of-the-art performance.

%% file: ms.bbl
\begin{thebibliography}{10}\itemsep=-1pt

\bibitem{CINM}
L.~Bao, B.~Wu, and W.~Liu.
\newblock Cnn in mrf: Video object segmentation via inference in a cnn-based
  higher-order spatio-temporal mrf.
\newblock In {\em Proceedings of the IEEE Conference on Computer Vision and
  Pattern Recognition}, pages 5977--5986, 2018.

\bibitem{OSVOS}
S.~Caelles, K.-K. Maninis, J.~Pont-Tuset, L.~Leal-Taix{\'e}, D.~Cremers, and
  L.~Van~Gool.
\newblock One-shot video object segmentation.
\newblock In {\em CVPR 2017}. IEEE, 2017.

\bibitem{chen2018deeplab}
L.-C. Chen, G.~Papandreou, I.~Kokkinos, K.~Murphy, and A.~L. Yuille.
\newblock Deeplab: Semantic image segmentation with deep convolutional nets,
  atrous convolution, and fully connected crfs.
\newblock {\em IEEE transactions on pattern analysis and machine intelligence},
  40(4):834--848, 2018.

\bibitem{PML}
Y.~Chen, J.~Pont-Tuset, A.~Montes, and L.~Van~Gool.
\newblock Blazingly fast video object segmentation with pixel-wise metric
  learning.
\newblock In {\em Proceedings of the IEEE Conference on Computer Vision and
  Pattern Recognition}, pages 1189--1198, 2018.

\bibitem{FAVOS}
J.~Cheng, Y.-H. Tsai, W.-C. Hung, S.~Wang, and M.-H. Yang.
\newblock Fast and accurate online video object segmentation via tracking
  parts.
\newblock In {\em The IEEE Conference on Computer Vision and Pattern
  Recognition (CVPR)}, June 2018.

\bibitem{SFL}
J.~Cheng, Y.-H. Tsai, S.~Wang, and M.-H. Yang.
\newblock Segflow: Joint learning for video object segmentation and optical
  flow.
\newblock In {\em Computer Vision (ICCV), 2017 IEEE International Conference
  on}, pages 686--695. IEEE, 2017.

\bibitem{cheng2015msra10k}
M.~Cheng.
\newblock Msra10k database, 2015.

\bibitem{LSE}
H.~Ci, C.~Wang, and Y.~Wang.
\newblock Video object segmentation by learning location-sensitive embeddings.
\newblock In {\em Proceedings of the European Conference on Computer Vision
  (ECCV)}, pages 501--516, 2018.

\bibitem{everingham2015pascal}
M.~Everingham, S.~A. Eslami, L.~Van~Gool, C.~K. Williams, J.~Winn, and
  A.~Zisserman.
\newblock The pascal visual object classes challenge: A retrospective.
\newblock {\em International journal of computer vision}, 111(1):98--136, 2015.

\bibitem{he2016deep}
K.~He, X.~Zhang, S.~Ren, and J.~Sun.
\newblock Deep residual learning for image recognition.
\newblock In {\em Proceedings of the IEEE conference on computer vision and
  pattern recognition}, pages 770--778, 2016.

\bibitem{hochreiter1997lstm}
S.~Hochreiter and J.~Schmidhuber.
\newblock Long short-term memory.
\newblock {\em Neural computation}, 9(8):1735--1780, 1997.

\bibitem{MGCRN}
P.~Hu, G.~Wang, X.~Kong, J.~Kuen, and Y.-P. Tan.
\newblock Motion-guided cascaded refinement network for video object
  segmentation.
\newblock In {\em Proceedings of the IEEE Conference on Computer Vision and
  Pattern Recognition}, pages 1400--1409, 2018.

\bibitem{VM}
Y.-T. Hu, J.-B. Huang, and A.~G. Schwing.
\newblock Videomatch: Matching based video object segmentation.
\newblock In {\em Proceedings of the European Conference on Computer Vision
  (ECCV)}, pages 54--70, 2018.

\bibitem{VPN}
V.~Jampani, R.~Gadde, and P.~V. Gehler.
\newblock Video propagation networks.
\newblock In {\em Proc. CVPR}, volume~6, page~7, 2017.

\bibitem{CTN}
W.-D. Jang and C.-S. Kim.
\newblock Online video object segmentation via convolutional trident network.
\newblock In {\em CVPR}, volume~1, page~7, 2017.

\bibitem{LuT}
A.~Khoreva, R.~Benenson, E.~Ilg, T.~Brox, and B.~Schiele.
\newblock Lucid data dreaming for object tracking.
\newblock In {\em The 2017 DAVIS Challenge on Video Object Segmentation - CVPR
  Workshops}, 2017.

\bibitem{Alexnet}
A.~Krizhevsky, I.~Sutskever, and G.~E. Hinton.
\newblock Imagenet classification with deep convolutional neural networks.
\newblock In {\em NIPS}, 2012.

\bibitem{DyeNet}
X.~Li and C.~Change~Loy.
\newblock Video object segmentation with joint re-identification and
  attention-aware mask propagation.
\newblock In {\em The European Conference on Computer Vision (ECCV)}, September
  2018.

\bibitem{FCN}
J.~Long, E.~Shelhamer, and T.~Darrell.
\newblock Fully convolutional networks for semantic segmentation.
\newblock In {\em Proceedings of the IEEE conference on computer vision and
  pattern recognition}, pages 3431--3440, 2015.

\bibitem{PReMVOS}
J.~Luiten, P.~Voigtlaender, and B.~Leibe.
\newblock Premvos: Proposal-generation, refinement and merging for the davis
  challenge on video object segmentation 2018, 2018.

\bibitem{OSVOS-S}
K.~Maninis, S.~Caelles, Y.~Chen, J.~Pont-Tuset, L.~Leal-Taixe, D.~Cremers, and
  L.~Van~Gool.
\newblock Video object segmentation without temporal information.
\newblock {\em IEEE Transactions on Pattern Analysis and Machine Intelligence},
  2018.

\bibitem{pytorch}
A.~Paszke, S.~Gross, S.~Chintala, G.~Chanan, E.~Yang, Z.~DeVito, Z.~Lin,
  A.~Desmaison, L.~Antiga, and A.~Lerer.
\newblock Automatic differentiation in pytorch.
\newblock 2017.

\bibitem{MSK}
F.~Perazzi, A.~Khoreva, R.~Benenson, B.~Schiele, and A.~Sorkine-Hornung.
\newblock Learning video object segmentation from static images.
\newblock In {\em Computer Vision and Pattern Recognition}, volume~2, 2017.

\bibitem{DAVIS16}
F.~Perazzi, J.~Pont-Tuset, B.~McWilliams, L.~Van~Gool, M.~Gross, and
  A.~Sorkine-Hornung.
\newblock A benchmark dataset and evaluation methodology for video object
  segmentation.
\newblock In {\em Proceedings of the IEEE Conference on Computer Vision and
  Pattern Recognition}, pages 724--732, 2016.

\bibitem{pinheiro2016learning}
P.~O. Pinheiro, T.-Y. Lin, R.~Collobert, and P.~Doll{\'a}r.
\newblock Learning to refine object segments.
\newblock In {\em European Conference on Computer Vision}, pages 75--91.
  Springer, 2016.

\bibitem{DAVIS17}
J.~Pont-Tuset, F.~Perazzi, S.~Caelles, P.~Arbel\'aez, A.~Sorkine-Hornung, and
  L.~{Van Gool}.
\newblock The 2017 davis challenge on video object segmentation.
\newblock {\em arXiv:1704.00675}, 2017.

\bibitem{ILSVRC15}
O.~Russakovsky, J.~Deng, H.~Su, J.~Krause, S.~Satheesh, S.~Ma, Z.~Huang,
  A.~Karpathy, A.~Khosla, M.~Bernstein, A.~C. Berg, and L.~Fei-Fei.
\newblock {ImageNet Large Scale Visual Recognition Challenge}.
\newblock {\em IJCV}, pages 1--42, April 2015.

\bibitem{SimonyanICLR2015}
K.~Simonyan and A.~Zisserman.
\newblock Very deep convolutional networks for large-scale image recognition.
\newblock In {\em ICLR}, 2015.

\bibitem{OFL}
Y.-H. Tsai, M.-H. Yang, and M.~J. Black.
\newblock Video segmentation via object flow.
\newblock In {\em Proceedings of the IEEE Conference on Computer Vision and
  Pattern Recognition}, pages 3899--3908, 2016.

\bibitem{OnAVOS}
P.~Voigtlaender and B.~Leibe.
\newblock Online adaptation of convolutional neural networks for video object
  segmentation.
\newblock In {\em British Machine Vision Conference 2017, {BMVC} 2017, London,
  UK, September 4-7, 2017}, 2017.

\bibitem{RGMP}
S.~Wug~Oh, J.-Y. Lee, K.~Sunkavalli, and S.~Joo~Kim.
\newblock Fast video object segmentation by reference-guided mask propagation.
\newblock In {\em Proceedings of the IEEE Conference on Computer Vision and
  Pattern Recognition}, pages 7376--7385, 2018.

\bibitem{YTVOS}
N.~Xu, L.~Yang, Y.~Fan, J.~Yang, D.~Yue, Y.~Liang, B.~L. Price, S.~Cohen, and
  T.~S. Huang.
\newblock Youtube-vos: Sequence-to-sequence video object segmentation.
\newblock In {\em Computer Vision - {ECCV} 2018 - 15th European Conference,
  Munich, Germany, September 8-14, 2018, Proceedings, Part {V}}, pages
  603--619, 2018.

\bibitem{YTVOS-extra}
N.~Xu, L.~Yang, Y.~Fan, D.~Yue, Y.~Liang, J.~Yang, and T.~S. Huang.
\newblock Youtube-vos: {A} large-scale video object segmentation benchmark.
\newblock {\em CoRR}, abs/1809.03327, 2018.

\bibitem{OSMN}
L.~Yang, Y.~Wang, X.~Xiong, J.~Yang, and A.~K. Katsaggelos.
\newblock Efficient video object segmentation via network modulation.
\newblock In {\em The IEEE Conference on Computer Vision and Pattern
  Recognition (CVPR)}, June 2018.

\end{thebibliography}
